\documentclass[sn-mathphys,Numbered]{sn-jnl}

\usepackage{graphicx}%
\usepackage{hyperref}%
\usepackage{multirow}%
\usepackage{amsmath,amssymb,amsfonts}%
\usepackage{amsthm}%
\usepackage{multicol}
\usepackage{mathrsfs}%
\usepackage[title]{appendix}%
\usepackage{xcolor}%
\usepackage{textcomp}%
\usepackage{manyfoot}%
\usepackage{booktabs}%
\usepackage{lineno}
\usepackage{algorithm}%
\usepackage{algorithmicx}%
\usepackage{algpseudocode}%
\usepackage{listings}%
\usepackage{array}
\usepackage[acronym]{glossaries}

\begin{document}

\newacronym{psg}{PSG}{Polysomnography}
\newacronym{mcc}{MCC}{Matthews Correlation Coefficient}
\newacronym{ece}{ECE}{Expected Calibration Error}
\newacronym{mlp}{MLP}{Multilayer perceptron}
\newacronym{cnn}{CNN}{Convolutional neural network}
\newacronym{lstm}{LSTM}{Long Short-Term Memory}
\newacronym{mle}{MLE}{Maximum Likelihood Estimator}
\newacronym{kl}{KL}{Kullback–Leibler divergence}
\newacronym{eeg}{EEG}{Electroencephalography}
\newacronym{ecg}{ECG}{Electrocardiography}
\newacronym{emg}{EMG}{Electromyography}
\newacronym{ck}{CK}{Cole-Kripke}
\newacronym{mesa}{MESA}{Multi-Ethnic Study
of Atherosclerosis}
 
\title[Article Title]{Advancing sleep detection by modelling weak label sets: A novel weakly supervised learning approach}


\author*[1]{\fnm{Matthias} \sur{Boeker}}\email{matthias@simula.no}

\author[1]{\fnm{Vajira} \sur{Thambawita}}\email{vajira@simula.no}

\author[1]{\fnm{Michael} \sur{Riegler}}\email{michael@simula.no}

\author[1,2]{\fnm{Pål} \sur{Halvorsen}}\email{paalh@simula.no}

\author[2,1]{\fnm{Hugo L.} \sur{Hammer}}\email{hugo.hammer@oslomet.no}

\affil*[1]{\orgdiv{Holistic Systems}, \orgname{SimulaMet}, \orgaddress{\street{Pilestredet 52}, \city{Oslo}, \postcode{0167}, \country{Norway}}}

\affil[2]{\orgdiv{Department of Computer Science}, \orgname{Oslomet}, \orgaddress{\street{Pilestredet 35}, \city{Oslo}, \postcode{0166}, \country{Norway}}}

\abstract{Understanding sleep and activity patterns plays a crucial role in physical and mental health. This study introduces a novel approach for sleep detection using weakly supervised learning for scenarios where reliable ground truth labels are unavailable. The proposed method relies on a set of weak labels, derived from the predictions generated by conventional sleep detection algorithms. Introducing a novel approach, we suggest a novel generalised non-linear statistical model in which the number of weak sleep labels is modelled as outcome of a binomial distribution.
The probability of sleep in the binomial distribution is linked to the outcomes of neural networks trained to detect sleep based on actigraphy. We show that maximizing the likelihood function of the model, is equivalent to minimizing the soft cross-entropy loss. Additionally, we explored the use of the Brier score as a loss function for weak labels. The efficacy of the suggested modelling framework was demonstrated using the Multi-Ethnic Study of Atherosclerosis dataset. A \gls{lstm} trained on the soft cross-entropy outperformed conventional sleep detection algorithms, other neural network architectures and loss functions in accuracy and model calibration. This research not only advances sleep detection techniques in scenarios where ground truth data is scarce but also contributes to the broader field of weakly supervised learning by introducing innovative approach in modelling sets of weak labels.}





\maketitle

\section{Introduction}\label{sec:intro}
Wrist-worn actigraphy is a non-invasive and convenient method for assessing sleep over several days. Algorithms for sleep detection based on actigraphy have been continuously developed and improved \cite{cho2019deep, haghayegh2020deep, granovsky2018actigraphy, barouni2020ambulatory}. Its effectiveness in identifying sleep patterns, sleep disorders, and neurobehavioural disorders has been well documented in various studies \cite{sadeh2011role, ancoli2003role, lam2008addressing, grierson2016circadian,hori201624,difrancesco2022role}. Actigraphy serves to objectively assess sleep habits and circadian rhythm disorders and is a tool for clarifying and contrasting sleep and activity patterns in severe psychiatric disorders, including schizophrenia \cite{tahmasian2013clinical}.

Reliable deep learning methods for sleep detection, defined by their confident and accurate predictions, require accurate ground truth labels \cite{cho2019deep, haghayegh2020deep, granovsky2018actigraphy}. But what should we do when such labels are unavailable? In general, generating valuable ground truth labels can be costly and time consuming \cite{zhou2018brief}. In sleep research, the gold standard for sleep detection is \gls{psg}. \gls{psg} monitors the participants' neurological signals, which are assessed by experts with regard to sleep \cite{anderer2023overview, rundo2019polysomnography}. The error rates for labelling are around $15\%$, which is considered sufficient to be accepted \cite{penzel2022sleep}. Accordingly, the validation of sleep detection algorithms which work on non-invasive signals like heart rate or actigraphy is difficult. Therefore, our paper investigates the potential of weakly supervised learning for sleep detection.

The general rise of machine learning, especially in the health field, has also led to numerous proposals for machine learning and deep learning models for sleep detection based on actigraphy \cite{cho2019deep, granovsky2018actigraphy, sundararajan2021sleep}. However, conventional algorithms are still a popular choice and commercial devices operate on them. While ActiGraph\texttrademark\ uses the algorithm by Sadeh and Cole-Kripke in their software ActiLife, Camntech\texttrademark\ uses the algorithm by Oakley \cite{oakley1997validation, actigraphwebpage}. The Oakley algorithm uses a threshold to distinguish between sleep and wake states \cite{schoch2019actimetry}. Sadeh et al. applied a discriminant analysis and derive a linear function for sleep classification \cite{sadeh1994activity}. Cole-Kripke et al. used a regression analysis to derive a linear function for the classification of sleep \cite{roger1992automatic}. In a comparative study, Patterson et. al compare the conventional algorithms with more recent machine learning methods \cite{patterson202340}. They show that the conventional algorithms are still competitive solutions for the sleep detection task. Each of these algorithms has been validated and fitted on specific, small sample sizes. In applied research, the researchers should consider suitability of these algorithms when applied to their purpose and dataset.

We reviewed the sample populations of conventional sleep detection algorithms, such as Sadeh, Cole-Kripke, Oakley, and Sazonova in order to understand the range of demographics that the algorithms cover \cite{oakley1997validation, sazonov2004activity}. While Sadeh's algorithm is calibrated on data from 36 subjects, the Cole-Kripke algorithm relies on data from 41 subjects. Oakley's algorithm is based on a more limited dataset, with only 11 participants \cite{roger1992automatic, sadeh1994activity}. The Sadeh algorithm includes 20 young adults between 20 and 25 years with a good balance between male and female participants \cite{sadeh1994activity}. Oakley used only men between 23 and 67 years \cite{oakley1997validation}. Cole-Kripke is fitted based on 32 men and 9 women with a mean age of 50.2 $(\pm 14.7)$ years. Sazanov et al. used logistic regression to classify sleep of 27 infants \cite{sazonov2004activity}. 

Choosing one algorithm for a case study is challenging when the study's population greatly differs from those in the original algorithm tests. The narrow samples sizes and potential biases make the choice difficult. Furthermore, establishing a ground truth is often not feasible or too expensive in research studies. This raises two problems for applied researchers. The first problem is to validate existing algorithms without ground truth. The second problem is that there is no way to integrate your own algorithms such as deep learning models if no ground truth is available.

\begin{figure}[htb!]
    \centering
    \includegraphics[width=1\linewidth]{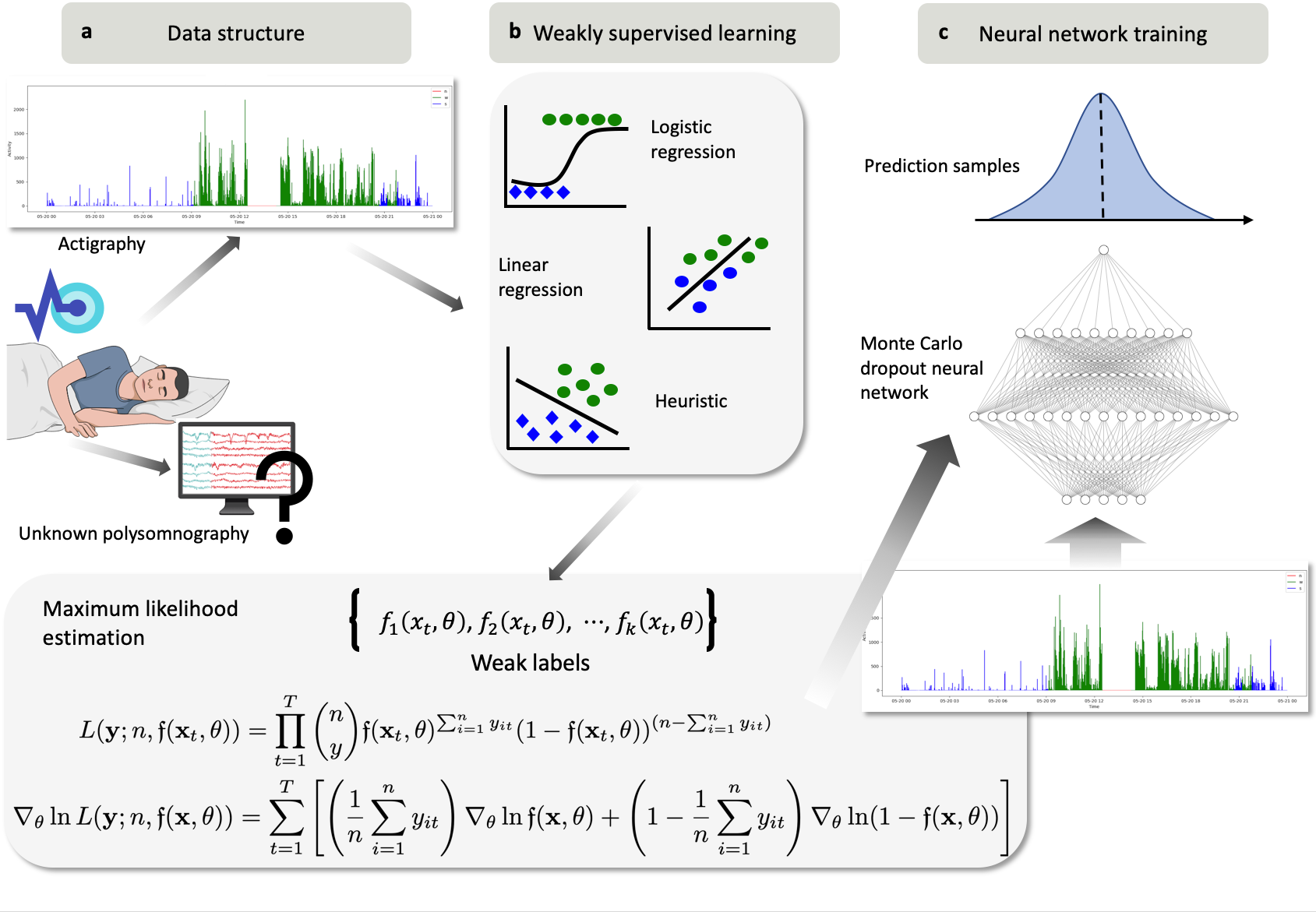}
    \caption{\textbf{An overview over the methodology of the work}. \textbf{a.} visualises the data structure of actigraphy and \gls{psg}. However, we tackle the problem of missing \gls{psg}, the ground truth for sleep. \textbf{b.} illustrates the idea behind the novel weakly supervised learning approach. An ensemble of conventional sleep detection algorithms like heuristics, linear or logistic regressions generate a set of weak labels. The number of weak labels is modelled as an outcome from a binomial distribution. We derived the soft cross-entropy loss function from the maximum likelihood estimation of this binomial distribution. In \textbf{c.} we show the training of Monte Carlo dropout neural networks based on actigraphy as input data and the weak labels. The neural network is trained with the derived soft cross-entropy loss. In the results, we show that the soft cross-entropy loss outperforms other loss functions in terms of classification performance and model calibration. The Monte Carlo dropout neural networks generate prediction samples instead of single value predictions. Hence, we gain insight into the prediction uncertainty, which is measured by the standard deviation of the prediction samples. }
    \label{fig:graphical_abstract}
\end{figure}


In this paper, we address the problem of detecting sleep from actigraphy when correctly annotated sleep is not available. Without labels for sleep, supervised machine learning, like previously successfully applied deep learning, is not possible. However, there are a number of simple and heuristic methods for sleep detection, which we called conventional sleep detection algorithms. We use the predictions of these algorithms to give us a set of labels. As the predictions do not represent the ground truth, we refer to them as weak labels. Essentially, we use an ensemble of conventional sleep detection algorithms to generate a set of weak labels. Figure~\ref{fig:graphical_abstract} illustrates the general intuition of the paper. 

It is not obvious how to learn from such multiple weak labels. Hence, we suggest a novel generalised non-linear statistical model where the number of weak labels predicting sleep at a given time point is modelled as an outcome from a binomial distribution. The probability of sleep in the binomial distribution is further linked to the outcome of neural networks trained to detect sleep on the basis of actigraphy. Since each weak label is the output of an algorithms, the labels are conditional probabilities. Accordingly, the labels are subject to uncertainty. To account for the inherent label uncertainty, we use neural networks with Monte Carlo dropout to approximate Bayesian inference \cite{gal2016dropout}. Monte Carlo dropout allows us to sample a large number of predictions instead of a single value. Their standard deviation gives us an understanding of the prediction uncertainty of our model.

We show that maximizing the binomial likelihood function of the suggested statistical model is equivalent to minimizing the soft cross-entropy loss. 
Our results demonstrate that a \gls{lstm} trained with the soft cross-entropy loss performs best in classification and model calibration, compared to the conventional algorithms, other neural network architectures and using other loss functions. While classification refers to the algorithm's ability to accurately identify sleep stages, calibration is crucial to ensure that the model's predictions are not only accurate, but also confident. Furthermore, we compare the standard deviation of the predictions of the Monte Carlo dropout neural networks with the standard deviation of the ensemble of the conventional algorithms. Both are an indicator of prediction uncertainty. We show that the prediction uncertainty in Monte Carlo dropout neural networks is lower compared to the uncertainty of the weak labels. 

In summary, the main contributions are as follows.
\begin{enumerate}
\item \textbf{In the Field of Sleep Research}:
\begin{enumerate}
\item We have developed an innovative weakly supervised approach for sleep analysis, tailored for situations where sleep ground truth is absent, and where conventional algorithms may not be adequately representative or specific.
\item We demonstrated the effectiveness of \gls{lstm} models, trained with soft cross-entropy loss, in enhancing both the accuracy and calibration of sleep detection, advancing the state of sleep analysis techniques.
\end{enumerate}
\item \textbf{In the Domain of Weakly Supervised Learning}:
\begin{enumerate}
    \item We have introduced a novel method that employs an ensemble of conventional sleep detection algorithms as a labelling mechanism, effectively bridging unsupervised and supervised learning paradigms.
    \item We have adapted the soft cross-entropy loss function directly from the binomial probability of a set of weak labels.
    \item We showcased the benefits of dropout neural networks in reducing prediction uncertainty, a significant improvement over conventional methods.
\end{enumerate}
\end{enumerate}
\section{Methods}\label{methods}
Our method tackles the problem of lacking ground truth or uncertain ground truth in sleep detection. We use an ensemble of conventional sleep detection algorithms to predict sleep and use the predictions as a set of weak labels. We propose a novel generalised non-linear statistical model where the number of weak labels predicting sleep at a given time point is modelled as an outcome from a binomial distribution. The probability of sleep in the binomial distribution is linked to actigraphy measurements using advanced deep neural networks. We show that maximizing the likelihood function of the model, is equivalent to minimizing the soft cross-entropy loss. The analysis is further cast into a Bayesian framework using dropout neural networks~\cite{gal2016dropout}. 

\subsection{Data}
The \gls{mesa} investigates the prevalence and progression of cardiovascular diseases. The study consists of actigraphy measures, sleep questionnaires and \gls{psg}. A total of 2,060 participants yielded successful \gls{psg} data, while 2,156 provided data through actigraphy, and 2,240 individuals completed sleep-related questionnaires \cite{chen2015racial}. A total of 1,836 participants successfully conducted the \gls{psg} and recorded their motor activity. The participants wore the Actiwatch Spectrum wrist actigraph (Philips Respironics, Murrysville, PA) for 7 consecutive days. The motor activity was aggregated to 30 second epochs \cite{chen2015racial}. \gls{psg} was performed using a 15-channel CompuMedics Somte system (Compumedics Ltd., Abbotsville, Australia). The recording system included electroencephalography \gls{eeg}, bilateral electrooculograms, chin electromyography \gls{emg}, bipolar electrocardiography \gls{ecg}, respiratory induction plethysmography for thoracic and abdominal movements, airflow monitoring via thermocouple and nasal pressure cannula, finger pulse oximetry and sensors for recording bilateral limb movements \cite{chen2015racial}. The \gls{psg} was annotated by experts into two states, sleep and wake. The analysis in our work only considered the experts annotations.

\subsection{Weakly supervised learning}\label{sec:inaccurate_supervised_learning}
Weakly supervised learning addresses the challenge of noisy labels \cite{zhou2018brief}. In our case, the weak labels are a set of predictions of an ensemble of $k$ conventional sleep algorithms. We consider the conventional algorithms as stochastic black boxes, whose outcomes are realisations of conditional distributions \cite{donmez2010unsupervised}. Each model $i \in \{1, \ldots, k\}$ produces a binary outcome point $y_{it} \in \{0,1\}$, where $1$ and $0$ indicate sleep and wake at a given time point $t$, respectively. The ensemble combines a series of conventional algorithms to obtain a series of weak labels that together form a stronger label.Bonab et al. show that when assembling the ensemble, not the number of models but their diversity is important \cite{bonab2019less}. Kuncheva et al. prove connections between accuracy and diversity in the ensemble in theoretical cases \cite{kuncheva2003measures}. In Section~\ref{sec:intro}, we discussed the generalisability of the conventional algorithms. The conventional algorithms are fitted on rather small datasets which under-represent certain demographic groups. Regarding the diversity, all algorithms are fitted on different datasets targeting different demographic groups. Thus, there is strong diversity in their ensemble, which allows for a smaller number of ensemble algorithms \cite{bonab2019less}. The applied conventional algorithms are Sadeh \cite{sadeh1994activity}, Oakley \cite{oakley1997validation}, Hidden Markov model \cite{li2020novel}, Cole-Kripke \cite{roger1992automatic}, and Sazonov \cite{sazonov2004activity}. 

\subsection{Statistical model}\label{sec:mathematical_model}
Given the labels from the conventional sleep algorithms we set up a statistical model and estimation procedure to find a best possible estimate for sleep at a given time point. Each model $i \in \{1, \ldots, k\}$ produces a binary outcome point $y_{it} \in \{0,1\}$, where $1$ and $0$ indicate sleep and wake at a given time point $t$, respectively. 
Let $y_t$ represent the number of  sleep algorithms that predicted that the participant slept at time $t$, i.e $y_t = \sum_{i=1}^k y_{it}$. We assume that $y_t$ is an outcome from a binomial distribution with parameters $n$ and $p_t$, where $p_t$ refers to the probability that an arbitrary sleep algorithm predicted that the participant slept at time $t$

\begin{equation}\label{eq:binominal_rv}
    y_t = \sum_{i=1}^k y_{it} \sim \text{Bin}(k, p_t)
\end{equation}
The probability $p_t$ is linked to the motor activity data form the $h$ previous time steps, $\mathbf{x_t} = x_t, \dots x_{t-h}$, using a neural network
\begin{equation}\label{eq:nn_approximation}
    p_t = \mathfrak{f} (\mathbf{x}_t, \theta)
\end{equation}
where $\mathfrak{f}$ refers to the neural network, and $\theta$ the parameters (weights) of the network. If a linear regression model were used instead of the suggested neural network, our suggested model would be equivalent to the commonly used generalised linear regression models. However, given the unstructured properties of the input data, we decided to use a neural network.

\subsection{Estimation}\label{sec:estimation}
The likelihood function for estimating the unknown parameters $\theta$, for the model presented above, is given as
\begin{equation}\label{eq:binomial_likelihood}
    L(\mathbf{y}; k, \mathfrak{f} (\mathbf{x}_t, \theta)) = \prod_{t=1}^T \binom{k}{y} \mathfrak{f} (\mathbf{x}_t, \theta)^{\sum_{i=1}^k y_{it}} (1 - \mathfrak{f} (\mathbf{x}_t, \theta))^{(k-\sum_{i=1}^n y_{it})}
\end{equation}
where $T$ refers to the length of the observation period and $\mathbf{y} = y_1, \ldots, y_T$. The log-likelihood function is further proportional to
\begin{align}\label{eq:log-likelihood}
\ln L(\mathbf{y}; k, \mathfrak{f} (\mathbf{x}_t, &\theta)) \propto \sum_{t=1}^T \left[ \left( \sum_{i=1}^k y_{it} \right) \ln \mathfrak{f} (\mathbf{x}_t, \theta) + \left( k-\sum_{i=1}^k y_{it} \right) \ln (1 - \mathfrak{f} (\mathbf{x}_t, \theta)) \right]\\
&\propto \sum_{t=1}^T \left[ \left( \frac{1}{k}\sum_{i=1}^k y_{it} \right) \ln \mathfrak{f} (\mathbf{x}_t, \theta) + \left(1-\frac{1}{k}\sum_{i=1}^k y_{it} \right) \ln (1 - \mathfrak{f} (\mathbf{x}_t, \theta)) \right] \\\label{eq:ce-similar}
&\propto \sum_{t=1}^T \left[ \hat{p}_t \ln \mathfrak{f} (\mathbf{x}_t, \theta) + \left(1-\hat{p}_t \right) \ln (1 - \mathfrak{f} (\mathbf{x}_t, \theta)) \right]
\end{align}
where $\hat{p}_t = \frac{1}{n}\sum_{i=1}^n y_{it}$, i.e. the average of the predictions in the ensemble, indicating how much the ensemble members agreed. The log likelihood function is equivalent to the commonly used cross entropy loss in machine learning \cite{wang2020comprehensive}. The derived cross-entropy loss measures the distance between the average of labels in the ensemble, $\hat{p}_t$, and the output of the neural network $\mathfrak{f} (\mathbf{x}, \theta))$. A full derivation of the Equation \ref{eq:ce-similar} from Equation \ref{eq:binomial_likelihood} is attached in Appendix \ref{ap:proof}. 

The choice of an appropriate loss function is crucial for optimizing the performance of the model. It not only measures the distance between our model's predictions and the ground truth, but also aligns with the probabilistic nature of our ensemble's outputs.

From the log-likelihood function above, we see that information from the ensemble are used through the average, $\hat{p}_t$. We also considered summarising the information in the ensemble using the majority voting \cite{sheng2008get}. The majority vote chooses the label with the most votes as overall labels. 
\begin{equation}\label{eq:majority_vote}
      y_{t} = \begin{cases} 1 & \text{if } \sum_{i=1}^{k} I(y_{it} > 0.5) > \frac{k}{2} \\
  0 & \text{otherwise}
  \end{cases} 
  \end{equation}
where $I(A)$ is the indicator function that returns one if $A$ is true, and else zero. I.e. that the majority vote is an indicator function which is one when the sum over the ensemble is greater than half it's size. The majority vote is a non-probabilistic and robust measure for aggregating labels. Majority voting gives us a binary outcome $\{0,1\}$ which can be interpreted as sleep or wake. However, the majority vote follows the all-or-nothing principle and eliminates information in how much the algorithms agreed among each other. With a binary ensemble label, we loose the information if $95\%$ or $51\%$ of the labels agreed. 
Labels aggregated by the majority vote are often called hard, those aggregated by the weighted average are called soft labels. 

The statistical modelling above derived at the soft cross-entropy loss, but we also applied the Brier score loss function that potentially could result in an efficient label prediction model. 
The Brier score is not a common loss function but mainly used as a proper scoring rule. Proper scoring rules provide a measure on how well the probabilities predict a class \cite{gneiting2007strictly}. The Brier score is the mean squared error of the probability and binary observations \cite{brier1950verification}. 
\begin{equation}\label{eq:brier_score}
     \mathcal{BS} = \frac{1}{T} \sum_{t=1}^T \left( \mathfrak{f} (\mathbf{x}, \theta) - y_t \right)^2
\end{equation}
So intuitively, the Brier score penalises models which predict very far from the actual observation and it matters how close the prediction is to the actual outcome. Accordingly, the score is very low when the predicted probability is close to the class value. For example the probability $0.1$ and class 0 are less penalised than the probability $0.45$ and class 0, even though the model would predict correct in both cases.
  
To quantify uncertainty, we applied the Monte Carlo dropout technique when training the neural networks \cite{kingma2015variational}. Gal and Ghahramani have shown that any neural network with dropout before each layer approximates a deep Gaussian process. The model draws binary numbers $d_i$ for each layer from a Bernoulli distributed random variable with probability $p_{\text{dropout}}$. A layer's output is set to zero when $d_i = 0$ \cite{gal2016dropout}.

Accordingly, the loss function with $L_2$ regularisation and dropout layers effectively minimises \Gls{kl} between approximated distribution and the posterior distribution of a deep Gaussian process \cite{gal2016dropout}. While complex posterior distributions are typically computed using Markov chain Monte Carlo methods \cite{giudici2013wiley}, in this context, the Monte Carlo dropout offers a more computationally feasible alternative. It allows us to maintain the numerical integrity of the training process and retain the derived soft cross-entropy loss, efficiently quantifying prediction uncertainty. Alternatively, variational inference could be considered to approximate the posterior distribution by minimising \gls{kl} between a variational and the true posterior distribution \cite{fox2012tutorial}.

\subsection{Neural Networks}
In this section, we describe the different used for $\mathfrak{f} (\mathbf{x}, \theta)$ in the model above.
We compared \gls{mlp}, \gls{cnn} and \gls{lstm} as neural network architectures to capture the nature of motor activity data. \gls{mlp} serves as a baseline architecture for the more complex \gls{cnn} and \gls{lstm}. \gls{mlp} works well on smaller datasets and is easy to train. But, the \gls{mlp} lacks the structure to understand sequential data. Through the convolutional layers, the \gls{cnn} is able to find local patterns in sequential data \cite{lecun2015deep}. This is of advantage in sleep detection, especially for wake periods during the night or short naps during the day. The recurrent architecture of the \gls{lstm} is developed for sequential data, especially to model long-time dependencies \cite{hochreiter1997long}. The \gls{lstm} is a natural choice for time series. However, \gls{cnn} and \gls{lstm} are computationally demanding models and require sufficiently large datasets for training.
All three architecture use batch normalisation after each fully connected layer. Batch normalisation stabilises the learning process and reduces the training steps \cite{ioffe2015batch}. Given the increased computational demands introduced by the Monte Carlo dropout, we have opted for efficient architectures in other aspects of the model design.

As mentioned in Section~\ref{sec:mathematical_model}, we used a dropout layer after each fully connected layer to approximate Bayesian inference \cite{gal2016dropout}. The dropout rate, $p_{\text{dropout}}$, was set to 0.5. The last layer of the neural networks is a sigmoid function with temperature scaling. Temperature scaling divides the final sigmoid layer by the $\tau$. With $\tau > 0$, the scaling increases the entropy of sigmoid function \cite{guo2017calibration}. Such temperature scaling is known to improve the calibration of the network, preventing it from becoming overly confident in its predictions.
The input into the neural networks is a time vector over the last $h=20$ time points, $\textbf{x}_t = x_t, x_{t-1}, \dots, x_{t-20}$, corresponding to a ten minute time interval. The output of the neural network is the number of ensemble members that predicted sleep, $y_t$, or equivalently the class probability $p_t = y_t/k$. The networks were implemented in Pytorch and the Adam optimiser was used for training \cite{paszke2017automatic}. The $L2$ penalty term $\lambda$ was set to $10^{-4}$, the learning rate was $10^{-5}$.

\subsection{Evaluation metrics}\label{sec:evaluation-metrics}
We look at two types of evaluation metrics, calibration and classification metrics, which have different purposes in model assessment. While classification metrics provide insights into the predictive power and accuracy of our models, calibration metrics help us grasp the confidence and reliability of their predictions. We will focus on introducing calibration metrics, while directing readers to overview literature on classification metrics in the medical field, as detailed in \cite{hicks2022evaluation, maier2206metrics}. 

The \gls{ece} represents the trade-off between accuracy and confidence. The predictions of the model are grouped in $M$ equally sized bins $B_m$.
\begin{equation}\label{eq:ece}
    ECE = \frac{|B_m|}{n} \sum_{m=1}^M |acc(B_m) - conf(B_m)|
\end{equation}
The \gls{ece} is a weighted average of the absolute differences between accuracy and confidence across multiple prediction bins \cite{naeini2015obtaining}. $conf(B_m)$ is the average predicted probability for the bin $B_m$. $acc(B_m)$ is the fraction of correctly predicted samples in a bin $B_m$. The discrepancy between accuracy and confidence is the calibration gap \cite{guo2017calibration}.

Shannon's entropy is a concept from the information theory. Given a set of possible events with certain probabilities it measures how much information is contained in this set \cite{shannon1948mathematical}, and given by 
\begin{equation}\label{eq:entropy}
    H = - \sum p_i log(p_i)
\end{equation}
A uniform distribution has a high entropy because it contains little information on whether an event will happen. Regarding the model calibration, entropy serves as a metric to assess the confidence level of a model's predictions, with higher entropy indicating lower prediction confidence \cite{braverman2020calibration, frenay2013classification}.

When modelling real-world data, there are many sources of uncertainty. We categorise these various sources in epistemic and aleatoric uncertainty \cite{der2009aleatory}. Epistemic uncertainty is uncertainty introduced by the model and can be reduced by the sufficient choice of the model and sufficient data. The aleatoric uncertainty is inherent in the stochasticity of the data and can be further divided in homoscedasticity and heteroscedasticity. While homoscedasticity assumes that the aleatoric uncertainty is constant over the data space, heteroscedasticity corresponds to varying noise of the data space \cite{seitzer2022pitfalls}. The dropout neural networks models homoscedastic noise \cite{venkatesh2019heteroscedastic}. If the model would account for heteroscedasticity or there is no heteroscedasticity in the stochasticity of the data, the prediction sample variance would be uniform randomly distributed.  
We estimate the prediction uncertainty of the ensemble through the sample variance across the ensemble's labels. The ensemble sample variance is given by
\begin{equation}\label{eq:ensemble_sample_variance}
\text{Var}(y_t)  = \frac{1}{k-1}\sum_{i=1}^k (y_{it} - \hat{p}_{t})^2
\end{equation}
The prediction uncertainty of the dropout neural networks is calculated by the sample variance over the number of sampled predictions. 

Any pattern in the prediction variance indicates heteroscedasticity. To examine the existence of heteroscedasticity, we calculate the average daily course of the prediction standard deviation for all subjects. Then, we fit a piece-wise polynomial function of degree 4 to the average daily course of the prediction. Based on this function, we calculate the intercept and the amplitude of the function. The intercept indicates the amount of uncertainty. We introduce the amplitude as a measure of pattern in the uncertainty. Based on these two measure, it is easier for us to compare how well the models reduce heteroscedasticity. 

\section{Results}\label{results}
In our experiments, we evaluate the classification performance and model calibration of the Monte Carlo dropout neural networks trained with weak labels. We compare three different neural network architectures, namely \gls{mlp}, \gls{cnn}, \gls{lstm}. Each of the neural network architecture is trained with three different loss functions. We compare the soft cross-entropy loss, which we derived from maximising the binomial likelihood of weak labels, with the hard cross-entropy and the Brier score loss. Additionally, we assess the performance of conventional sleep detection algorithms and investigate whether neural networks perform better than the conventional algorithms with which they are trained. Finally, we compare the magnitude and structure of the prediction uncertainty of the neural networks with the uncertainty of the conventional algorithm ensemble. 

The experiments are based on the \gls{mesa} dataset. The data contain one night per participant with \gls{psg} as ground truth \cite{chen2015racial, zhang2018national}. The data is divided into a training, validation and test set. Only the test set contains the ground truth, the training and validation sets contain the weak labels. The models are trained and evaluated per person. The results presented in this section are the mean and standard deviation of the individual performance measures for the 1833 participants in the MESA study.

Table~\ref{tab:conventional_algorithm_results} showed the classification results of the conventional algorithms and their ensemble. Notably, only Sazanov's algorithm completely fails to detect sleep. The ensemble achieves the highest \gls{mcc} however, the results were very close to each other considering the standard deviation. While, the ensemble scores the highest accuracy, the Oakley algorithms achieves a lower standard deviation. Generally, the Oakley algorithms performs best, achieving the best F1 score, Cohen's kappa and specificity. The results confirm that the ensemble of the conventional algorithms achieved at least a comparable performance compared to each individual algorithm. Moreover, the ensemble maintains always a comparable small standard deviation. 
Overall, the conventional algorithms performed well in detecting sleep. However, when in doubt about the applicability of a single algorithm, the ensemble of algorithms managed to establish a solid baseline.
\begin{table}[htb!]
\centering
\label{tab:conventional_algorithm_results}
\begin{tabular}{|l|l|l|l|l|l|l|}
\hline
Metric & CK & Ensemble & HMM & Oakley & Sadeh & Sazonov \\
\hline
Accuracy & $0.761 \pm 0.079$ & $\textbf{0.786} \pm 0.073$ & $0.678 \pm 0.076$ & $0.787 \pm 0.077$ & $0.773 \pm 0.070$ & $0.500 \pm 0.000$ \\
F1 & $0.697 \pm 0.118$ & $0.718 \pm 0.116$ & $0.623 \pm 0.111$ & $\textbf{0.720} \pm 0.117$ & $0.704 \pm 0.120$ & $0.000 \pm 0.000$ \\
Kappa & $0.451 \pm 0.175$ & $0.500 \pm 0.168$ & $0.295 \pm 0.145$ & $\textbf{0.507} \pm 0.175$ & $0.471 \pm 0.159$ & $0.000 \pm 0.000$ \\
MCC & $0.525 \pm 0.147$ & $\textbf{0.563} \pm 0.143$ & $0.380 \pm 0.148$ & $0.561 \pm 0.149$ & $0.543 \pm 0.137$ & $0.000 \pm 0.000$ \\
Sensitivity & $\textbf{0.976} \pm 0.031$ & $0.968 \pm 0.052$ & $0.934 \pm 0.119$ & $0.950 \pm 0.047$ & $0.973 \pm 0.078$ & $0.000 \pm 0.000$ \\
Specificity & $0.546 \pm 0.155$ & $0.604 \pm 0.138$ & $0.421 \pm 0.135$ & $0.624 \pm 0.148$ & $0.573 \pm 0.130$ & $\textbf{1.000} \pm 0.000$ \\
\hline
\end{tabular}
\caption{\textbf{The classification performance of the conventional algorithms.} The algorithms are evaluated based on one night of annotated \gls{psg}. CK refers to the Cole-Kripke algorithm.}
\end{table}

Table~\ref{tab:nn_loss_function_comparision} summarised the performance of three different neural network architectures and the loss functions.
The classification performance metrics will show us which neural network architecture works best on the data. Furthermore, we examine whether the Brier score, the soft cross-entropy or the hard label cross-entropy work better in terms of model calibration.
The \gls{lstm} outperformed the other neural network architectures with respect to accuracy, \gls{mcc}, F1 score and Cohen's kappa. However, the other two neural network architecture still performed better than all conventional algorithms. Moreover, the soft cross-entropy loss function performed better in these metrics compared to the other loss functions, especially for the \gls{cnn} and \gls{lstm}. 
\begin{table}[htb!]
\label{tab:nn_loss_function_comparision}
\newcolumntype{L}{>{\raggedright\arraybackslash}p{0.7cm}} 
\begin{tabular}{|L|l|l|l|l|}
\hline
NN Type & Metric & Brier Score & Cross-entropy & Soft cross-entropy \\
\hline
 & Accuracy & $0.787 \pm 0.082$ & $0.798 \pm 0.077$ & $0.806 \pm 0.086$ \\
 & ECE & $0.236 \pm 0.099$ & $0.228 \pm 0.096$ & $0.207 \pm 0.096$ \\
 & Entropy & $0.121 \pm 0.058$ & $0.084 \pm 0.049$ & $0.083 \pm 0.056$ \\
 CNN & F1 & $0.718 \pm 0.125$ & $0.729 \pm 0.121$ & $0.737 \pm 0.130$ \\
 & Kappa & $0.505 \pm 0.182$ & $0.528 \pm 0.176$ & $0.551 \pm 0.190$ \\
 & MCC & $0.561 \pm 0.162$ & $0.578 \pm 0.156$ & $0.590 \pm 0.175$ \\
 & Sensitivity & $0.953 \pm 0.081$ & $0.949 \pm 0.071$ & $0.925 \pm 0.099$ \\
 & Specificity & $0.622 \pm 0.136$ & $0.647 \pm 0.129$ & $0.688 \pm 0.126$ \\
 & Variance & $0.010 \pm 0.008$ & $0.009 \pm 0.006$ & $0.010 \pm 0.007$ \\
\hline
 & Accuracy & $0.8057 \pm 0.0751$ & $0.7973 \pm 0.0717$ & $0.8233 \pm 0.0766$ \\
 & ECE & $0.1930 \pm 0.0935$ & $0.2341 \pm 0.0965$ & $0.1910 \pm 0.0934$ \\
 & Entropy & $0.2112 \pm 0.0949$ & $0.0788 \pm 0.0435$ & $0.0656 \pm 0.0663$ \\
LSTM & F1 Score & $0.7342 \pm 0.1239$ & $0.7253 \pm 0.1216$ & $0.7538 \pm 0.1239$ \\
 & Kappa & $0.5425 \pm 0.1765$ & $0.5214 \pm 0.1700$ & $0.5852 \pm 0.1786$ \\
 & MCC & $0.5900 \pm 0.1549$ & $0.5766 \pm 0.1478$ & $0.6197 \pm 0.1599$ \\
 & Sensitivity & $0.9493 \pm 0.0590$ & $0.9622 \pm 0.0442$ & $0.9307 \pm 0.0635$ \\
 & Specificity & $0.6621 \pm 0.1276$ & $0.6325 \pm 0.1225$ & $0.7158 \pm 0.1228$ \\
 & Variance & $0.0084 \pm 0.0112$ & $0.0034 \pm 0.0051$ & $0.0076 \pm 0.0136$ \\
\hline
 & Accuracy & $0.7940 \pm 0.0710$ & $0.7938 \pm 0.0709$ & $0.7939 \pm 0.0710$ \\
 & ECE & $0.2389 \pm 0.0962$ & $0.2392 \pm 0.0963$ & $0.2390 \pm 0.0963$ \\
 & Entropy & $0.0902 \pm 0.0436$ & $0.0903 \pm 0.0437$ & $0.0902 \pm 0.0436$ \\
MLP & F1 Score & $0.7246 \pm 0.1161$ & $0.7244 \pm 0.1160$ & $0.7245 \pm 0.1161$ \\
 & Kappa & $0.5152 \pm 0.1663$ & $0.5148 \pm 0.1661$ & $0.5151 \pm 0.1662$ \\
 & MCC & $0.5731 \pm 0.1433$ & $0.5727 \pm 0.1431$ & $0.5730 \pm 0.1433$ \\
 & Sensitivity & $0.9655 \pm 0.0436$ & $0.9656 \pm 0.0435$ & $0.9654 \pm 0.0437$ \\
 & Specificity & $0.6225 \pm 0.1252$ & $0.6220 \pm 0.1251$ & $0.6224 \pm 0.1252$ \\
 & Variance & $0.0069 \pm 0.0052$ & $0.0069 \pm 0.0052$ & $0.0069 \pm 0.0052$ \\
\hline
\end{tabular}
\caption{\label{tab:comparative_architecture}\textbf{Comparison of neural network performance metrics for different loss functions.} The table presents the performance metrics of three neural network types when trained with Brier Score, cross-entropy, and soft cross-entropy. Each metric is accompanied by its mean and standard deviation, providing a comprehensive overview of the networks' predictive performances and calibrations.}
\end{table}
Additional to the well-known classification performance metrics, we examined metrics of model calibration. The metrics used for model calibration are the expected calibration error, the entropy over the predictions and, the variance of the Monte Carlo dropout prediction sampling. The variance is a measure of the model prediction uncertainty. The entropy of the predicted probabilities serves as an indicator of the model's predictive certainty. A high entropy value in a neural network's output, indicating a uniform distribution, suggests output probabilities close to  $0.5\%$. Conversely, a low entropy value indicates confident predictions, usually with probabilities near $0.9\%$ or $0.1\%$. The \gls{lstm} with soft cross-entropy loss achieved the lowest entropy. While the entropy measures the confidence of our model, \gls{ece} indicates how well the predicted probabilities match the actual outcomes \cite{posocco2021estimating}. The \gls{lstm} trained with the soft-cross entropy scores the lowest \gls{ece}. Generally, the \gls{lstm} scored the lowest \gls{ece} across all three loss functions. This means that the \gls{lstm} is the best calibrated neural network architecure. The variance is nearly the same for \gls{cnn} and \gls{mlp}. Only the \gls{lstm} has a much lower variance with the hard cross-entropy loss function.

\begin{figure}[htb!]
    \centering
    \includegraphics[trim=35mm 0mm 40mm 0mm, clip,width=1\linewidth]{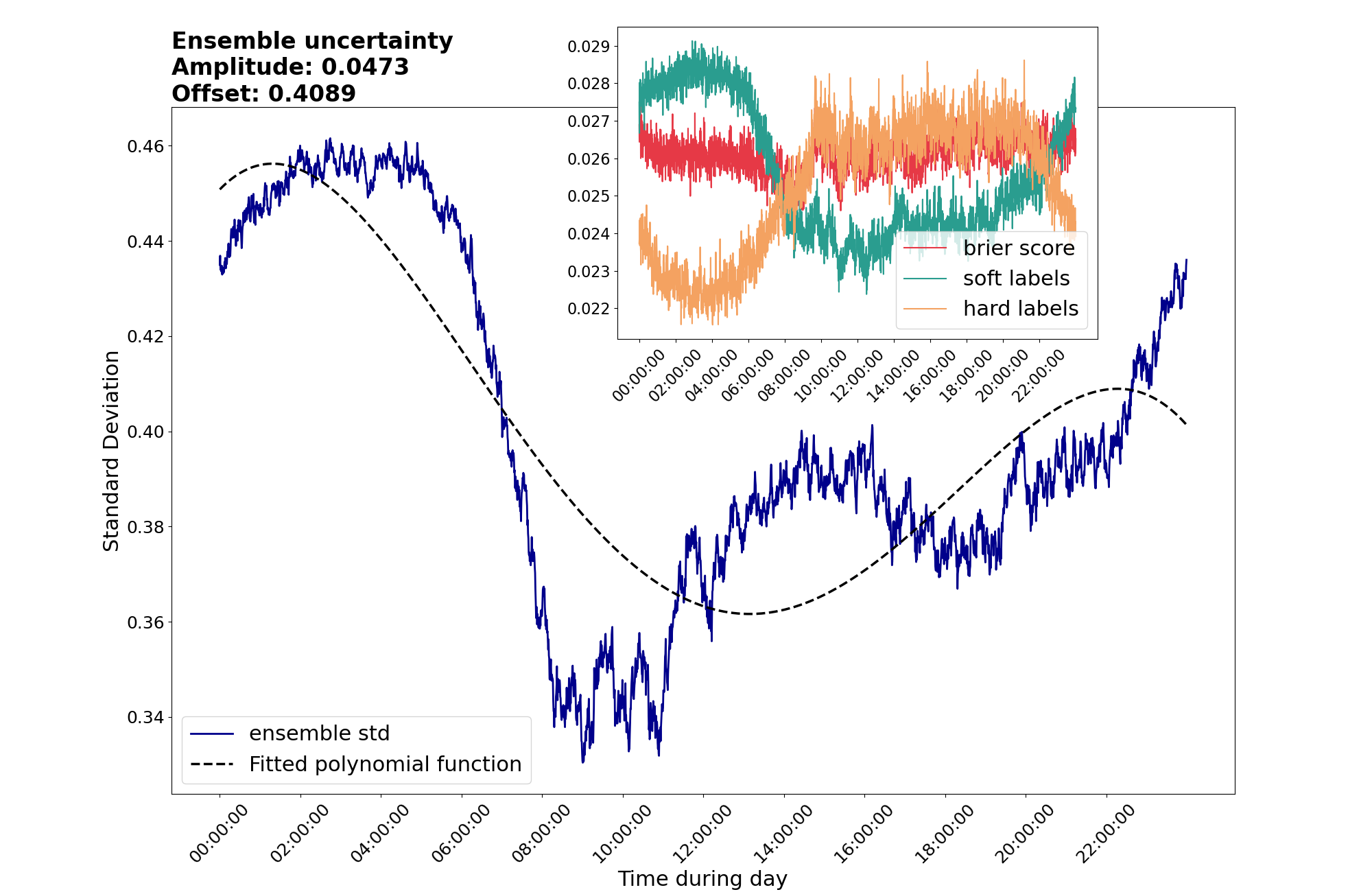}
    \caption{\textbf{The comparison of the time dependent prediction uncertainty of the ensemble model with the \gls{lstm} trained with three different loss functions}. The uncertainty is averaged over 204 subjects over one day. The main diagram highlights the prediction uncertainty of the ensemble with a blue line, while the inset diagram provides a detailed view of the significantly smaller prediction uncertainties of the \gls{lstm} models for better visualisation. Patterns and structures in uncertainty are indicators of heteroscedasticity.  Additionally, the main plot features a dashed line, fitting a cubic curve with an amplitude of 0.0473 and an offset of 0.4089. This cubic spline serves to model the heteroscedasticity of the prediction uncertainty.}
    \label{fig:ensemble_uncertainty_comparision}
\end{figure}

Figure~\ref{fig:ensemble_uncertainty_comparision} shows the standard deviation of the ensemble in the large plot and compares it to the standard deviation of the sampled prediction of the Monte Carlo dropout neural networks. The prediction standard deviation is averaged for each 30 seconds of the day and averaged over 216 participants. The \gls{mesa} dataset contains a variety of sleep conditions. To get balanced data, we selected 108 participants with insomnia and sampled 108 without any sleep condition. The standard deviation of the Monte Carlo dropout neural network predictions and the ensemble indicate the prediction uncertainty. 
A structure of prediction uncertainty indicates heteroscedasticity. Analysing heteroscedasticity helps us to understand the structure of predition uncertainty. This understanding improves model assessment and trustworthiness. We fitted a fourth-degree piece-wise polynomial function to approximate the heteroscedasticity structure. The polynomial function is the dashed line in Figure~\ref{fig:ensemble_uncertainty_comparision}. The first important observation is the difference in scaling, the uncertainty of the ensemble is much bigger than the one of the Monte Carlo dropout neural networks. The uncertainty shows a daily pattern which indicates that the uncertainty of predicting sleep is highest during the night. 

\begin{figure}[htb!]
    \centering
    \includegraphics[width=1\linewidth]{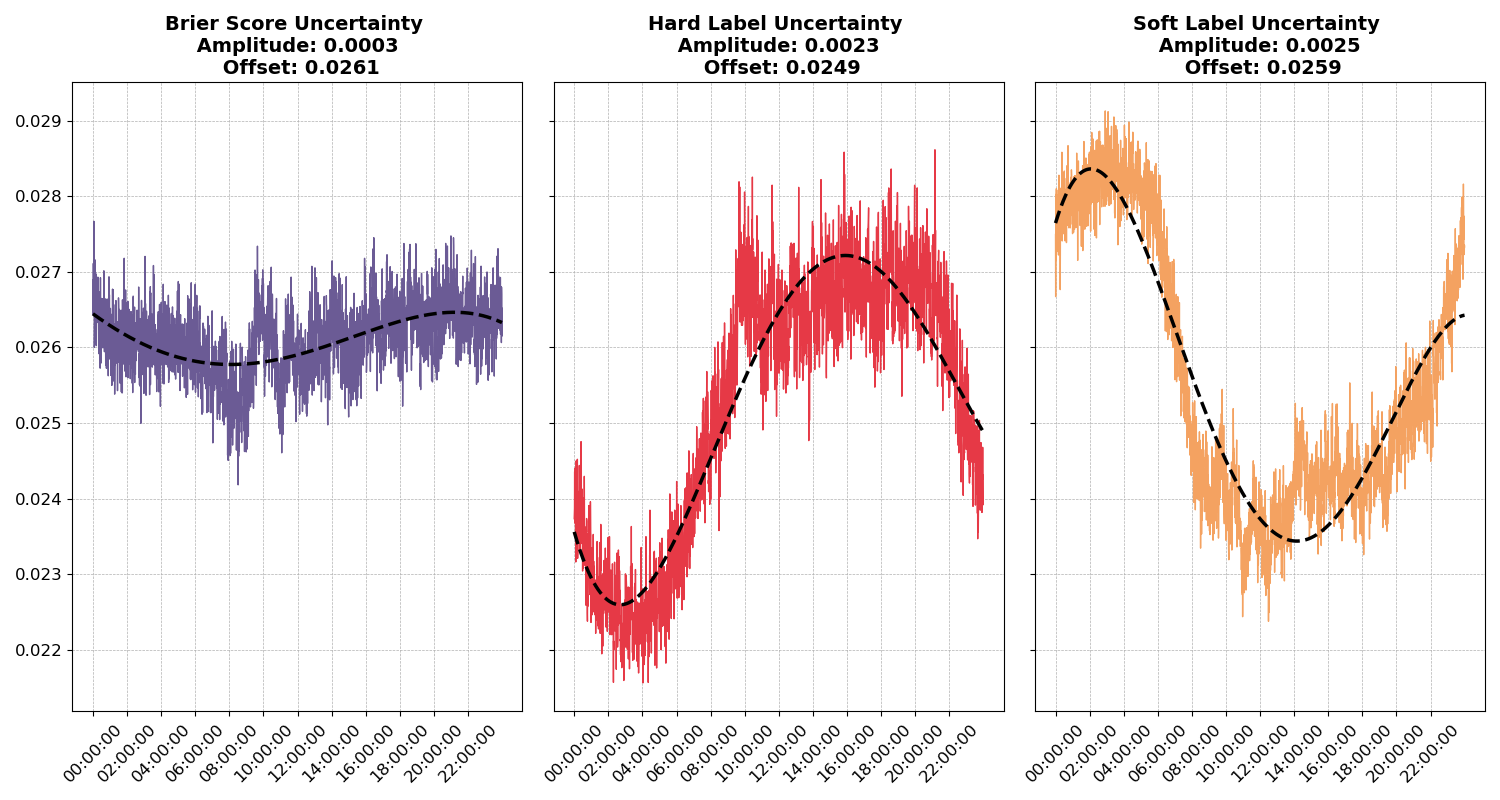}
    \caption{\textbf{Detailed comparison of the time dependent prediction uncertainty of the three loss functions.} The figure compares the prediction uncertainty in size and shape of the \gls{lstm} trained with the three different loss functions. The figure is a detailed presentation of the inset plot of Figure~\ref{fig:ensemble_uncertainty_comparision}. Patterns and structures in uncertainty are indicators of heteroscedasticity. The dashed lines represent fitted cubic splines. The offset indicates the size of the uncertainty. The amplitude shows the structure.}
    \label{fig:loss_function_uncertainty_comparision}
\end{figure}

Figure~\ref{fig:loss_function_uncertainty_comparision} compares the prediction uncertainty of Monte Carlo dropout neural networks trained with three different loss functions: Brier score, soft label cross-entropy, and hard label cross-entropy. The Brier score displays the least structured uncertainty, while the soft label cross-entropy shows a pattern similar to the ensemble uncertainty. The soft and hard label cross-entropy produce two different structures of uncertainty. While the soft label uncertainty resembles in structure the ensemble uncertainty, the hard label cross-entropy generates an opposite structure. The hard label uncertainty seems to be less certain during the night then during the day. We conclude that the Brier score results in the least heteroscedasticity, as evidenced by the amplitude of the polynomial function being $30^{-4}$. Overall, our proposed method of using neural networks trained on the basis of ensemble predictions considerably reduces uncertainty.

We reproduced the results on heteroscedasticity on the Depresjon and Psykose dataset, which have been merged \cite{garcia2018depresjon, jakobsen2020psykose}. The results showed that the \gls{lstm} reduces the magnitude of uncertainty. The details about the datasets and the results are attached in Appendix \ref{sec:A1Data}. 

\section{Discussion}\label{sec:discussion}
Our work presents a novel method for sleep detection using weakly supervised learning. We provide a solution to the need for validation in situations where \gls{psg} is unaffordable or in general ground truth is unavailable. We demonstrate that the \gls{lstm} trained with soft cross-entropy loss performs best compared to conventional algorithms, the ensemble of conventional algorithms and other deep learning architectures. Additionally, our work offers an incremental contribution to the field of weakly supervised learning by presenting a derivation of the soft cross-entropy loss from the binomial likelihood of weak labels. 

Our results are similar to the findings of Patterson et al. \cite{patterson202340} who validate conventional algorithms and deep learning architectures on the \gls{mesa}. They showed that the \gls{lstm} performs best compared to other methods. However, Patterson et al. used \gls{psg} as labels for training. This limits their application of deep learning to ground truth datasets. Our weakly trained \gls{lstm} and \gls{cnn} showed comparable performance metrics results, without the access to ground truth. And, we demonstrated the superiority of deep learning methods trained on weak labels generated in an unsupervised setup over conventional algorithms. Moreover, the \gls{lstm} trained with our derived soft cross-entropy loss is the best calibrated model compared to other architectures and loss functions. The soft cross-entropy loss produces the most confident and accurate models. 
Our approach enables sleep researchers to apply deep learning algorithms without the need of ground truth. Furthermore, when applying dropout neural networks, researchers can validate the models based on prediction uncertainty. However, our study shows limitations in handling heteroscedastic prediction uncertainty. The uncertainty revealed a temporal structure. The Brier score revealed a smoother uncertainty profiles, but exhibited poorer calibration. This is further evidenced by the comparative results in Table~\ref{tab:comparative_architecture}.
This is due the mean squared nature of the loss function which penalises the error uniformly.

In conclusion, deep learning models trained on weak labels can match the performance of those trained on ground truth data \cite{patterson202340}. This significant achievement not only aligns with but also extends the current benchmarks in sleep research, underscoring the potential of our approach in model calibration and uncertainty management.

\section{Limitations and future research}
Selecting appropriate models for labelling is crucial, especially in actigraphy-based sleep detection where conventional algorithms often involve heuristic, linear, or logistic functions fitted on different case study datasets \cite{sadeh1994activity, oakley1997validation, roger1992automatic}. Choosing more computationally intensive methods in the ensemble could significantly increase the overall complexity. Furthermore, relying solely on labelling algorithms fitted to the specific dataset in question may compromise the generalisability of the approach. In addition, the computational intensity deep learning models, especially for dropouts, can pose a challenge. Therefore, we suggest that the ensemble of conventional algorithms itself can improve the practice of sleep researchers.

Sleep research focuses often on the intersection of sleep problems and mental health. Mental health conditions like depression and schizophrenia can disturb the circadian rhythms \cite{jakobsen2020psykose, garcia2018depresjon}. We applied our method on the large sample population of the \gls{mesa} and two additional datasets concerning mental health. However our study has not extensively explored the model's calibration and prediction uncertainty across diverse groups with sleep and mental health problems. Since the circadian rhythms are often disturbed, the sleep algorithms might not calibrate as well or even discriminate between groups. Future work could investigate this and explore potential transfer learning approaches. 
 
The core idea of generating a set of weak labels can be transferred to various domains, such as human annotation in label crowdsourcing, image object recognition and clinical applications, as in \cite{bakker2023scoring, zhou2018brief, zheng2021uncertainty, reamaroon2018accounting, ju2022improving}. If the reliability of labels is in doubt, unsupervised model predictions can enhance the weak label set. While we currently use a simple average in our soft cross-entropy loss, adopting weighted averages based on the reliability of weak labels represents a further refinement. Finally, the uncertainty can be included into the loss function to get a uncertainty-aware model. This can reduce heteroscedasticity and calibrate the models even better to weak labelling.

\backmatter
\section{Data availability}
The \gls{mesa} study is available through Zhang et al. \cite{zhang2018national}. The Psykose dataset is openly available on this \hyperlink{https://datasets.simula.no/psykose/}{repository} \cite{jakobsen2020psykose}. The Depresjon dataset is openly available on this \hyperlink{https://datasets.simula.no/depresjon/}{repository} \cite{garcia2018depresjon}. 
\section{Code availability}
The Pyhton code is available on \hyperlink{https://github.com/matthiasboeker/ensemble_sleep_detection}{GitHub} \cite{boeker2023ensemble}. The repository contains the neural network training and implementation, visualisation and  experiment scripts.
\section{Authors' contributions}
MB was responsible for the conceptualisation of the project, the main idea, developing the methods and implementing the experiments and neural networks, and validating the work. Additionally, MB wrote the initial draft, reviewed and edited the text. HLH had a key role in developing the methodology, conceptualising, writing and reviewing the methods section. He was also involved in the review and editing of the overall text.  VT provided assistance in the early stages, contributing to conceptualization and experiment hardware setup. VT played a role in the review process. MR and PH contributed primarily through their efforts in reviewing the manuscript, offering valuable insights to improve the writing. 
\section{Competing interests}
The authors declare no competing interests.

\begin{multicols}{2}
\printglossary
\bibliography{bibliography}


\begin{thebibliography}{59}
\ifx \bisbn   \undefined \def \bisbn  #1{ISBN #1}\fi
\ifx \binits  \undefined \def \binits#1{#1}\fi
\ifx \bauthor  \undefined \def \bauthor#1{#1}\fi
\ifx \batitle  \undefined \def \batitle#1{#1}\fi
\ifx \bjtitle  \undefined \def \bjtitle#1{#1}\fi
\ifx \bvolume  \undefined \def \bvolume#1{\textbf{#1}}\fi
\ifx \byear  \undefined \def \byear#1{#1}\fi
\ifx \bissue  \undefined \def \bissue#1{#1}\fi
\ifx \bfpage  \undefined \def \bfpage#1{#1}\fi
\ifx \blpage  \undefined \def \blpage #1{#1}\fi
\ifx \burl  \undefined \def \burl#1{\textsf{#1}}\fi
\ifx \doiurl  \undefined \def \doiurl#1{\url{https://doi.org/#1}}\fi
\ifx \betal  \undefined \def \betal{\textit{et al.}}\fi
\ifx \binstitute  \undefined \def \binstitute#1{#1}\fi
\ifx \binstitutionaled  \undefined \def \binstitutionaled#1{#1}\fi
\ifx \bctitle  \undefined \def \bctitle#1{#1}\fi
\ifx \beditor  \undefined \def \beditor#1{#1}\fi
\ifx \bpublisher  \undefined \def \bpublisher#1{#1}\fi
\ifx \bbtitle  \undefined \def \bbtitle#1{#1}\fi
\ifx \bedition  \undefined \def \bedition#1{#1}\fi
\ifx \bseriesno  \undefined \def \bseriesno#1{#1}\fi
\ifx \blocation  \undefined \def \blocation#1{#1}\fi
\ifx \bsertitle  \undefined \def \bsertitle#1{#1}\fi
\ifx \bsnm \undefined \def \bsnm#1{#1}\fi
\ifx \bsuffix \undefined \def \bsuffix#1{#1}\fi
\ifx \bparticle \undefined \def \bparticle#1{#1}\fi
\ifx \barticle \undefined \def \barticle#1{#1}\fi
\bibcommenthead
\ifx \bconfdate \undefined \def \bconfdate #1{#1}\fi
\ifx \botherref \undefined \def \botherref #1{#1}\fi
\ifx \url \undefined \def \url#1{\textsf{#1}}\fi
\ifx \bchapter \undefined \def \bchapter#1{#1}\fi
\ifx \bbook \undefined \def \bbook#1{#1}\fi
\ifx \bcomment \undefined \def \bcomment#1{#1}\fi
\ifx \oauthor \undefined \def \oauthor#1{#1}\fi
\ifx \citeauthoryear \undefined \def \citeauthoryear#1{#1}\fi
\ifx \endbibitem  \undefined \def \endbibitem {}\fi
\ifx \bconflocation  \undefined \def \bconflocation#1{#1}\fi
\ifx \arxivurl  \undefined \def \arxivurl#1{\textsf{#1}}\fi
\csname PreBibitemsHook\endcsname

\bibitem[\protect\citeauthoryear{Cho et~al.}{2019}]{cho2019deep}
\begin{barticle}
\bauthor{\bsnm{Cho}, \binits{T.}},
\bauthor{\bsnm{Sunarya}, \binits{U.}},
\bauthor{\bsnm{Yeo}, \binits{M.}},
\bauthor{\bsnm{Hwang}, \binits{B.}},
\bauthor{\bsnm{Koo}, \binits{Y.S.}},
\bauthor{\bsnm{Park}, \binits{C.}}:
\batitle{Deep-actinet: End-to-end deep learning architecture for automatic sleep-wake detection using wrist actigraphy}.
\bjtitle{Electronics}
\bvolume{8}(\bissue{12}),
\bfpage{1461}
(\byear{2019})
\end{barticle}
\endbibitem

\bibitem[\protect\citeauthoryear{Haghayegh et~al.}{2020}]{haghayegh2020deep}
\begin{barticle}
\bauthor{\bsnm{Haghayegh}, \binits{S.}},
\bauthor{\bsnm{Khoshnevis}, \binits{S.}},
\bauthor{\bsnm{Smolensky}, \binits{M.H.}},
\bauthor{\bsnm{Diller}, \binits{K.R.}},
\bauthor{\bsnm{Castriotta}, \binits{R.J.}}:
\batitle{Deep neural network sleep scoring using combined motion and heart rate variability data}.
\bjtitle{Sensors}
\bvolume{21}(\bissue{1}),
\bfpage{25}
(\byear{2020})
\end{barticle}
\endbibitem

\bibitem[\protect\citeauthoryear{Granovsky et~al.}{2018}]{granovsky2018actigraphy}
\begin{botherref}
\oauthor{\bsnm{Granovsky}, \binits{L.}},
\oauthor{\bsnm{Shalev}, \binits{G.}},
\oauthor{\bsnm{Yacovzada}, \binits{N.}},
\oauthor{\bsnm{Frank}, \binits{Y.}},
\oauthor{\bsnm{Fine}, \binits{S.}}:
Actigraphy-based sleep/wake pattern detection using convolutional neural networks.
arXiv preprint arXiv:1802.07945
(2018)
\end{botherref}
\endbibitem

\bibitem[\protect\citeauthoryear{Barouni et~al.}{2020}]{barouni2020ambulatory}
\begin{barticle}
\bauthor{\bsnm{Barouni}, \binits{A.}},
\bauthor{\bsnm{Ottenbacher}, \binits{J.}},
\bauthor{\bsnm{Schneider}, \binits{J.}},
\bauthor{\bsnm{Feige}, \binits{B.}},
\bauthor{\bsnm{Riemann}, \binits{D.}},
\bauthor{\bsnm{Herlan}, \binits{A.}},
\bauthor{\bsnm{El~Hardouz}, \binits{D.}},
\bauthor{\bsnm{McLennan}, \binits{D.}}:
\batitle{Ambulatory sleep scoring using accelerometers—distinguishing between nonwear and sleep/wake states}.
\bjtitle{PeerJ}
\bvolume{8},
\bfpage{8284}
(\byear{2020})
\end{barticle}
\endbibitem

\bibitem[\protect\citeauthoryear{Sadeh}{2011}]{sadeh2011role}
\begin{barticle}
\bauthor{\bsnm{Sadeh}, \binits{A.}}:
\batitle{The role and validity of actigraphy in sleep medicine: an update}.
\bjtitle{Sleep medicine reviews}
\bvolume{15}(\bissue{4}),
\bfpage{259}--\blpage{267}
(\byear{2011})
\end{barticle}
\endbibitem

\bibitem[\protect\citeauthoryear{Ancoli-Israel et~al.}{2003}]{ancoli2003role}
\begin{barticle}
\bauthor{\bsnm{Ancoli-Israel}, \binits{S.}},
\bauthor{\bsnm{Cole}, \binits{R.}},
\bauthor{\bsnm{Alessi}, \binits{C.}},
\bauthor{\bsnm{Chambers}, \binits{M.}},
\bauthor{\bsnm{Moorcroft}, \binits{W.}},
\bauthor{\bsnm{Pollak}, \binits{C.P.}}:
\batitle{The role of actigraphy in the study of sleep and circadian rhythms}.
\bjtitle{Sleep}
\bvolume{26}(\bissue{3}),
\bfpage{342}--\blpage{392}
(\byear{2003})
\end{barticle}
\endbibitem

\bibitem[\protect\citeauthoryear{Lam}{2008}]{lam2008addressing}
\begin{barticle}
\bauthor{\bsnm{Lam}, \binits{R.}}:
\batitle{Addressing circadian rhythm disturbances in depressed patients}.
\bjtitle{Journal of Psychopharmacology}
\bvolume{22}(\bissue{7\_suppl}),
\bfpage{13}--\blpage{18}
(\byear{2008})
\end{barticle}
\endbibitem

\bibitem[\protect\citeauthoryear{Grierson et~al.}{2016}]{grierson2016circadian}
\begin{barticle}
\bauthor{\bsnm{Grierson}, \binits{A.B.}},
\bauthor{\bsnm{Hickie}, \binits{I.B.}},
\bauthor{\bsnm{Naismith}, \binits{S.L.}},
\bauthor{\bsnm{Hermens}, \binits{D.F.}},
\bauthor{\bsnm{Scott}, \binits{E.M.}},
\bauthor{\bsnm{Scott}, \binits{J.}}:
\batitle{Circadian rhythmicity in emerging mood disorders: state or trait marker?}
\bjtitle{International journal of bipolar disorders}
\bvolume{4}(\bissue{1}),
\bfpage{1}--\blpage{7}
(\byear{2016})
\end{barticle}
\endbibitem

\bibitem[\protect\citeauthoryear{Hori et~al.}{2016}]{hori201624}
\begin{barticle}
\bauthor{\bsnm{Hori}, \binits{H.}},
\bauthor{\bsnm{Koga}, \binits{N.}},
\bauthor{\bsnm{Hidese}, \binits{S.}},
\bauthor{\bsnm{Nagashima}, \binits{A.}},
\bauthor{\bsnm{Kim}, \binits{Y.}},
\bauthor{\bsnm{Higuchi}, \binits{T.}},
\bauthor{\bsnm{Kunugi}, \binits{H.}}:
\batitle{24-h activity rhythm and sleep in depressed outpatients}.
\bjtitle{Journal of psychiatric research}
\bvolume{77},
\bfpage{27}--\blpage{34}
(\byear{2016})
\end{barticle}
\endbibitem

\bibitem[\protect\citeauthoryear{Difrancesco et~al.}{2022}]{difrancesco2022role}
\begin{barticle}
\bauthor{\bsnm{Difrancesco}, \binits{S.}},
\bauthor{\bsnm{Penninx}, \binits{B.W.}},
\bauthor{\bsnm{Riese}, \binits{H.}},
\bauthor{\bsnm{Giltay}, \binits{E.J.}},
\bauthor{\bsnm{Lamers}, \binits{F.}}:
\batitle{The role of depressive symptoms and symptom dimensions in actigraphy-assessed sleep, circadian rhythm, and physical activity}.
\bjtitle{Psychological Medicine}
\bvolume{52}(\bissue{13}),
\bfpage{2760}--\blpage{2766}
(\byear{2022})
\end{barticle}
\endbibitem

\bibitem[\protect\citeauthoryear{Tahmasian et~al.}{2013}]{tahmasian2013clinical}
\begin{barticle}
\bauthor{\bsnm{Tahmasian}, \binits{M.}},
\bauthor{\bsnm{Khazaie}, \binits{H.}},
\bauthor{\bsnm{Golshani}, \binits{S.}},
\bauthor{\bsnm{Avis}, \binits{K.T.}}:
\batitle{Clinical application of actigraphy in psychotic disorders: a systematic review}.
\bjtitle{Current psychiatry reports}
\bvolume{15},
\bfpage{1}--\blpage{15}
(\byear{2013})
\end{barticle}
\endbibitem

\bibitem[\protect\citeauthoryear{Zhou}{2018}]{zhou2018brief}
\begin{barticle}
\bauthor{\bsnm{Zhou}, \binits{Z.-H.}}:
\batitle{A brief introduction to weakly supervised learning}.
\bjtitle{National science review}
\bvolume{5}(\bissue{1}),
\bfpage{44}--\blpage{53}
(\byear{2018})
\end{barticle}
\endbibitem

\bibitem[\protect\citeauthoryear{Anderer et~al.}{2023}]{anderer2023overview}
\begin{barticle}
\bauthor{\bsnm{Anderer}, \binits{P.}},
\bauthor{\bsnm{Ross}, \binits{M.}},
\bauthor{\bsnm{Cerny}, \binits{A.}},
\bauthor{\bsnm{Vasko}, \binits{R.}},
\bauthor{\bsnm{Shaw}, \binits{E.}},
\bauthor{\bsnm{Fonseca}, \binits{P.}}:
\batitle{Overview of the hypnodensity approach to scoring sleep for polysomnography and home sleep testing}.
\bjtitle{Frontiers in Sleep}
\bvolume{2},
\bfpage{1163477}
(\byear{2023})
\end{barticle}
\endbibitem

\bibitem[\protect\citeauthoryear{Rundo and Downey~III}{2019}]{rundo2019polysomnography}
\begin{barticle}
\bauthor{\bsnm{Rundo}, \binits{J.V.}},
\bauthor{\bsnm{Downey~III}, \binits{R.}}:
\batitle{Polysomnography}.
\bjtitle{Handbook of clinical neurology}
\bvolume{160},
\bfpage{381}--\blpage{392}
(\byear{2019})
\end{barticle}
\endbibitem

\bibitem[\protect\citeauthoryear{Penzel}{2022}]{penzel2022sleep}
\begin{botherref}
\oauthor{\bsnm{Penzel}, \binits{T.}}:
Sleep scoring moving from visual scoring towards automated scoring.
Oxford University Press US
(2022)
\end{botherref}
\endbibitem

\bibitem[\protect\citeauthoryear{Sundararajan et~al.}{2021}]{sundararajan2021sleep}
\begin{barticle}
\bauthor{\bsnm{Sundararajan}, \binits{K.}},
\bauthor{\bsnm{Georgievska}, \binits{S.}},
\bauthor{\bsnm{Te~Lindert}, \binits{B.H.}},
\bauthor{\bsnm{Gehrman}, \binits{P.R.}},
\bauthor{\bsnm{Ramautar}, \binits{J.}},
\bauthor{\bsnm{Mazzotti}, \binits{D.R.}},
\bauthor{\bsnm{Sabia}, \binits{S.}},
\bauthor{\bsnm{Weedon}, \binits{M.N.}},
\bauthor{\bsnm{Someren}, \binits{E.J.}},
\bauthor{\bsnm{Ridder}, \binits{L.}}, \betal:
\batitle{Sleep classification from wrist-worn accelerometer data using random forests}.
\bjtitle{Scientific reports}
\bvolume{11}(\bissue{1}),
\bfpage{24}
(\byear{2021})
\end{barticle}
\endbibitem

\bibitem[\protect\citeauthoryear{Oakley}{1997}]{oakley1997validation}
\begin{barticle}
\bauthor{\bsnm{Oakley}, \binits{N.R.}}:
\batitle{Validation with polysomnography of the sleepwatch sleep/wake scoring algorithm used by the actiwatch activity monitoring system. mini mitter co}.
\bjtitle{Sleep}
\bvolume{2},
\bfpage{0}--\blpage{140}
(\byear{1997})
\end{barticle}
\endbibitem

\bibitem[\protect\citeauthoryear{Actilife}{}]{actigraphwebpage}
\begin{botherref}
\oauthor{\bsnm{Actilife}}:
Actilife - Detect Sleep Periods.
\url{https://actigraphcorp.my.site.com/support/s/article/What-does-the-Detect-Sleep-Periods-button-do-and-how-does-it-work}.
Accessed on October 30, 2023
\end{botherref}
\endbibitem

\bibitem[\protect\citeauthoryear{Schoch et~al.}{2019}]{schoch2019actimetry}
\begin{barticle}
\bauthor{\bsnm{Schoch}, \binits{S.F.}},
\bauthor{\bsnm{Jenni}, \binits{O.G.}},
\bauthor{\bsnm{Kohler}, \binits{M.}},
\bauthor{\bsnm{Kurth}, \binits{S.}}:
\batitle{Actimetry in infant sleep research: an approach to facilitate comparability}.
\bjtitle{Sleep}
\bvolume{42}(\bissue{7}),
\bfpage{083}
(\byear{2019})
\end{barticle}
\endbibitem

\bibitem[\protect\citeauthoryear{Sadeh et~al.}{1994}]{sadeh1994activity}
\begin{barticle}
\bauthor{\bsnm{Sadeh}, \binits{A.}},
\bauthor{\bsnm{Sharkey}, \binits{M.}},
\bauthor{\bsnm{Carskadon}, \binits{M.A.}}:
\batitle{Activity-based sleep-wake identification: an empirical test of methodological issues}.
\bjtitle{Sleep}
\bvolume{17}(\bissue{3}),
\bfpage{201}--\blpage{207}
(\byear{1994})
\end{barticle}
\endbibitem

\bibitem[\protect\citeauthoryear{Roger}{1992}]{roger1992automatic}
\begin{barticle}
\bauthor{\bsnm{Roger}, \binits{J.}}:
\batitle{Automatic sleep/wake identification from wrist activity}.
\bjtitle{Sleep}
\bvolume{15}(\bissue{5}),
\bfpage{461}--\blpage{469}
(\byear{1992})
\end{barticle}
\endbibitem

\bibitem[\protect\citeauthoryear{Patterson et~al.}{2023}]{patterson202340}
\begin{barticle}
\bauthor{\bsnm{Patterson}, \binits{M.R.}},
\bauthor{\bsnm{Nunes}, \binits{A.A.}},
\bauthor{\bsnm{Gerstel}, \binits{D.}},
\bauthor{\bsnm{Pilkar}, \binits{R.}},
\bauthor{\bsnm{Guthrie}, \binits{T.}},
\bauthor{\bsnm{Neishabouri}, \binits{A.}},
\bauthor{\bsnm{Guo}, \binits{C.C.}}:
\batitle{40 years of actigraphy in sleep medicine and current state of the art algorithms}.
\bjtitle{NPJ Digital Medicine}
\bvolume{6}(\bissue{1}),
\bfpage{51}
(\byear{2023})
\end{barticle}
\endbibitem

\bibitem[\protect\citeauthoryear{Sazonov et~al.}{2004}]{sazonov2004activity}
\begin{barticle}
\bauthor{\bsnm{Sazonov}, \binits{E.}},
\bauthor{\bsnm{Sazonova}, \binits{N.}},
\bauthor{\bsnm{Schuckers}, \binits{S.}},
\bauthor{\bsnm{Neuman}, \binits{M.}},
\bauthor{\bsnm{Group}, \binits{C.S.}}, \betal:
\batitle{Activity-based sleep--wake identification in infants}.
\bjtitle{Physiological measurement}
\bvolume{25}(\bissue{5}),
\bfpage{1291}
(\byear{2004})
\end{barticle}
\endbibitem

\bibitem[\protect\citeauthoryear{Gal and Ghahramani}{2016}]{gal2016dropout}
\begin{bchapter}
\bauthor{\bsnm{Gal}, \binits{Y.}},
\bauthor{\bsnm{Ghahramani}, \binits{Z.}}:
\bctitle{Dropout as a bayesian approximation: Representing model uncertainty in deep learning}.
In: \bbtitle{International Conference on Machine Learning},
pp. \bfpage{1050}--\blpage{1059}
(\byear{2016}).
\bcomment{PMLR}
\end{bchapter}
\endbibitem

\bibitem[\protect\citeauthoryear{Chen et~al.}{2015}]{chen2015racial}
\begin{barticle}
\bauthor{\bsnm{Chen}, \binits{X.}},
\bauthor{\bsnm{Wang}, \binits{R.}},
\bauthor{\bsnm{Zee}, \binits{P.}},
\bauthor{\bsnm{Lutsey}, \binits{P.L.}},
\bauthor{\bsnm{Javaheri}, \binits{S.}},
\bauthor{\bsnm{Alc{\'a}ntara}, \binits{C.}},
\bauthor{\bsnm{Jackson}, \binits{C.L.}},
\bauthor{\bsnm{Williams}, \binits{M.A.}},
\bauthor{\bsnm{Redline}, \binits{S.}}:
\batitle{Racial/ethnic differences in sleep disturbances: the multi-ethnic study of atherosclerosis (mesa)}.
\bjtitle{Sleep}
\bvolume{38}(\bissue{6}),
\bfpage{877}--\blpage{888}
(\byear{2015})
\end{barticle}
\endbibitem

\bibitem[\protect\citeauthoryear{Donmez et~al.}{2010}]{donmez2010unsupervised}
\begin{botherref}
\oauthor{\bsnm{Donmez}, \binits{P.}},
\oauthor{\bsnm{Lebanon}, \binits{G.}},
\oauthor{\bsnm{Balasubramanian}, \binits{K.}}:
Unsupervised supervised learning i: Estimating classification and regression errors without labels.
Journal of Machine Learning Research
\textbf{11}(4)
(2010)
\end{botherref}
\endbibitem

\bibitem[\protect\citeauthoryear{Bonab and Can}{2019}]{bonab2019less}
\begin{barticle}
\bauthor{\bsnm{Bonab}, \binits{H.}},
\bauthor{\bsnm{Can}, \binits{F.}}:
\batitle{Less is more: A comprehensive framework for the number of components of ensemble classifiers}.
\bjtitle{IEEE Transactions on neural networks and learning systems}
\bvolume{30}(\bissue{9}),
\bfpage{2735}--\blpage{2745}
(\byear{2019})
\end{barticle}
\endbibitem

\bibitem[\protect\citeauthoryear{Kuncheva and Whitaker}{2003}]{kuncheva2003measures}
\begin{barticle}
\bauthor{\bsnm{Kuncheva}, \binits{L.I.}},
\bauthor{\bsnm{Whitaker}, \binits{C.J.}}:
\batitle{Measures of diversity in classifier ensembles and their relationship with the ensemble accuracy}.
\bjtitle{Machine learning}
\bvolume{51},
\bfpage{181}--\blpage{207}
(\byear{2003})
\end{barticle}
\endbibitem

\bibitem[\protect\citeauthoryear{Li et~al.}{2020}]{li2020novel}
\begin{barticle}
\bauthor{\bsnm{Li}, \binits{X.}},
\bauthor{\bsnm{Zhang}, \binits{Y.}},
\bauthor{\bsnm{Jiang}, \binits{F.}},
\bauthor{\bsnm{Zhao}, \binits{H.}}:
\batitle{A novel machine learning unsupervised algorithm for sleep/wake identification using actigraphy}.
\bjtitle{Chronobiology international}
\bvolume{37}(\bissue{7}),
\bfpage{1002}--\blpage{1015}
(\byear{2020})
\end{barticle}
\endbibitem

\bibitem[\protect\citeauthoryear{Wang et~al.}{2020}]{wang2020comprehensive}
\begin{botherref}
\oauthor{\bsnm{Wang}, \binits{Q.}},
\oauthor{\bsnm{Ma}, \binits{Y.}},
\oauthor{\bsnm{Zhao}, \binits{K.}},
\oauthor{\bsnm{Tian}, \binits{Y.}}:
A comprehensive survey of loss functions in machine learning.
Annals of Data Science,
1--26
(2020)
\end{botherref}
\endbibitem

\bibitem[\protect\citeauthoryear{Sheng et~al.}{2008}]{sheng2008get}
\begin{bchapter}
\bauthor{\bsnm{Sheng}, \binits{V.S.}},
\bauthor{\bsnm{Provost}, \binits{F.}},
\bauthor{\bsnm{Ipeirotis}, \binits{P.G.}}:
\bctitle{Get another label? improving data quality and data mining using multiple, noisy labelers}.
In: \bbtitle{Proceedings of the 14th ACM SIGKDD International Conference on Knowledge Discovery and Data Mining},
pp. \bfpage{614}--\blpage{622}
(\byear{2008})
\end{bchapter}
\endbibitem

\bibitem[\protect\citeauthoryear{Gneiting and Raftery}{2007}]{gneiting2007strictly}
\begin{barticle}
\bauthor{\bsnm{Gneiting}, \binits{T.}},
\bauthor{\bsnm{Raftery}, \binits{A.E.}}:
\batitle{Strictly proper scoring rules, prediction, and estimation}.
\bjtitle{Journal of the American statistical Association}
\bvolume{102}(\bissue{477}),
\bfpage{359}--\blpage{378}
(\byear{2007})
\end{barticle}
\endbibitem

\bibitem[\protect\citeauthoryear{Brier}{1950}]{brier1950verification}
\begin{barticle}
\bauthor{\bsnm{Brier}, \binits{G.W.}}:
\batitle{Verification of forecasts expressed in terms of probability}.
\bjtitle{Monthly weather review}
\bvolume{78}(\bissue{1}),
\bfpage{1}--\blpage{3}
(\byear{1950})
\end{barticle}
\endbibitem

\bibitem[\protect\citeauthoryear{Kingma et~al.}{2015}]{kingma2015variational}
\begin{botherref}
\oauthor{\bsnm{Kingma}, \binits{D.P.}},
\oauthor{\bsnm{Salimans}, \binits{T.}},
\oauthor{\bsnm{Welling}, \binits{M.}}:
Variational dropout and the local reparameterization trick.
Advances in neural information processing systems
\textbf{28}
(2015)
\end{botherref}
\endbibitem

\bibitem[\protect\citeauthoryear{Giudici et~al.}{2013}]{giudici2013wiley}
\begin{bbook}
\bauthor{\bsnm{Giudici}, \binits{P.}},
\bauthor{\bsnm{Givens}, \binits{G.H.}},
\bauthor{\bsnm{Mallick}, \binits{B.K.}}:
\bbtitle{Wiley Series in Computational Statistics}
vol. \bseriesno{596}.
\bpublisher{Wiley Online Library}, \blocation{???}
(\byear{2013})
\end{bbook}
\endbibitem

\bibitem[\protect\citeauthoryear{Fox and Roberts}{2012}]{fox2012tutorial}
\begin{barticle}
\bauthor{\bsnm{Fox}, \binits{C.W.}},
\bauthor{\bsnm{Roberts}, \binits{S.J.}}:
\batitle{A tutorial on variational bayesian inference}.
\bjtitle{Artificial intelligence review}
\bvolume{38},
\bfpage{85}--\blpage{95}
(\byear{2012})
\end{barticle}
\endbibitem

\bibitem[\protect\citeauthoryear{LeCun et~al.}{2015}]{lecun2015deep}
\begin{barticle}
\bauthor{\bsnm{LeCun}, \binits{Y.}},
\bauthor{\bsnm{Bengio}, \binits{Y.}},
\bauthor{\bsnm{Hinton}, \binits{G.}}:
\batitle{Deep learning}.
\bjtitle{nature}
\bvolume{521}(\bissue{7553}),
\bfpage{436}--\blpage{444}
(\byear{2015})
\end{barticle}
\endbibitem

\bibitem[\protect\citeauthoryear{Hochreiter and Schmidhuber}{1997}]{hochreiter1997long}
\begin{barticle}
\bauthor{\bsnm{Hochreiter}, \binits{S.}},
\bauthor{\bsnm{Schmidhuber}, \binits{J.}}:
\batitle{Long short-term memory}.
\bjtitle{Neural computation}
\bvolume{9}(\bissue{8}),
\bfpage{1735}--\blpage{1780}
(\byear{1997})
\end{barticle}
\endbibitem

\bibitem[\protect\citeauthoryear{Ioffe and Szegedy}{2015}]{ioffe2015batch}
\begin{bchapter}
\bauthor{\bsnm{Ioffe}, \binits{S.}},
\bauthor{\bsnm{Szegedy}, \binits{C.}}:
\bctitle{Batch normalization: Accelerating deep network training by reducing internal covariate shift}.
In: \bbtitle{International Conference on Machine Learning},
pp. \bfpage{448}--\blpage{456}
(\byear{2015}).
\bcomment{pmlr}
\end{bchapter}
\endbibitem

\bibitem[\protect\citeauthoryear{Guo et~al.}{2017}]{guo2017calibration}
\begin{bchapter}
\bauthor{\bsnm{Guo}, \binits{C.}},
\bauthor{\bsnm{Pleiss}, \binits{G.}},
\bauthor{\bsnm{Sun}, \binits{Y.}},
\bauthor{\bsnm{Weinberger}, \binits{K.Q.}}:
\bctitle{On calibration of modern neural networks}.
In: \bbtitle{International Conference on Machine Learning},
pp. \bfpage{1321}--\blpage{1330}
(\byear{2017}).
\bcomment{PMLR}
\end{bchapter}
\endbibitem

\bibitem[\protect\citeauthoryear{Paszke et~al.}{2017}]{paszke2017automatic}
\begin{botherref}
\oauthor{\bsnm{Paszke}, \binits{A.}},
\oauthor{\bsnm{Gross}, \binits{S.}},
\oauthor{\bsnm{Chintala}, \binits{S.}},
\oauthor{\bsnm{Chanan}, \binits{G.}},
\oauthor{\bsnm{Yang}, \binits{E.}},
\oauthor{\bsnm{DeVito}, \binits{Z.}},
\oauthor{\bsnm{Lin}, \binits{Z.}},
\oauthor{\bsnm{Desmaison}, \binits{A.}},
\oauthor{\bsnm{Antiga}, \binits{L.}},
\oauthor{\bsnm{Lerer}, \binits{A.}}:
Automatic differentiation in pytorch
(2017)
\end{botherref}
\endbibitem

\bibitem[\protect\citeauthoryear{Hicks et~al.}{2022}]{hicks2022evaluation}
\begin{barticle}
\bauthor{\bsnm{Hicks}, \binits{S.A.}},
\bauthor{\bsnm{Str{\"u}mke}, \binits{I.}},
\bauthor{\bsnm{Thambawita}, \binits{V.}},
\bauthor{\bsnm{Hammou}, \binits{M.}},
\bauthor{\bsnm{Riegler}, \binits{M.A.}},
\bauthor{\bsnm{Halvorsen}, \binits{P.}},
\bauthor{\bsnm{Parasa}, \binits{S.}}:
\batitle{On evaluation metrics for medical applications of artificial intelligence}.
\bjtitle{Scientific reports}
\bvolume{12}(\bissue{1}),
\bfpage{5979}
(\byear{2022})
\end{barticle}
\endbibitem

\bibitem[\protect\citeauthoryear{Maier-Hein et~al.}{}]{maier2206metrics}
\begin{botherref}
\oauthor{\bsnm{Maier-Hein}, \binits{L.}},
\oauthor{\bsnm{Reinke}, \binits{A.}},
\oauthor{\bsnm{Godau}, \binits{P.}},
\oauthor{\bsnm{Tizabi}, \binits{M.}},
\oauthor{\bsnm{B{\"u}ttner}, \binits{F.}},
\oauthor{\bsnm{Christodoulou}, \binits{E.}},
\oauthor{\bsnm{Glocker}, \binits{B.}},
\oauthor{\bsnm{Isensee}, \binits{F.}},
\oauthor{\bsnm{Kleesiek}, \binits{J.}},
\oauthor{\bsnm{Kozubek}, \binits{M.}}, et al.:
Metrics reloaded: Recommendations for image analysis validation. arxiv 2023.
arXiv preprint arXiv:2206.01653
\end{botherref}
\endbibitem

\bibitem[\protect\citeauthoryear{Naeini et~al.}{2015}]{naeini2015obtaining}
\begin{bchapter}
\bauthor{\bsnm{Naeini}, \binits{M.P.}},
\bauthor{\bsnm{Cooper}, \binits{G.}},
\bauthor{\bsnm{Hauskrecht}, \binits{M.}}:
\bctitle{Obtaining well calibrated probabilities using bayesian binning}.
In: \bbtitle{Proceedings of the AAAI Conference on Artificial Intelligence},
vol. \bseriesno{29}
(\byear{2015})
\end{bchapter}
\endbibitem

\bibitem[\protect\citeauthoryear{Shannon}{1948}]{shannon1948mathematical}
\begin{barticle}
\bauthor{\bsnm{Shannon}, \binits{C.E.}}:
\batitle{A mathematical theory of communication}.
\bjtitle{The Bell system technical journal}
\bvolume{27}(\bissue{3}),
\bfpage{379}--\blpage{423}
(\byear{1948})
\end{barticle}
\endbibitem

\bibitem[\protect\citeauthoryear{Braverman et~al.}{2020}]{braverman2020calibration}
\begin{bchapter}
\bauthor{\bsnm{Braverman}, \binits{M.}},
\bauthor{\bsnm{Chen}, \binits{X.}},
\bauthor{\bsnm{Kakade}, \binits{S.}},
\bauthor{\bsnm{Narasimhan}, \binits{K.}},
\bauthor{\bsnm{Zhang}, \binits{C.}},
\bauthor{\bsnm{Zhang}, \binits{Y.}}:
\bctitle{Calibration, entropy rates, and memory in language models}.
In: \bbtitle{International Conference on Machine Learning},
pp. \bfpage{1089}--\blpage{1099}
(\byear{2020}).
\bcomment{PMLR}
\end{bchapter}
\endbibitem

\bibitem[\protect\citeauthoryear{Fr{\'e}nay and Verleysen}{2013}]{frenay2013classification}
\begin{barticle}
\bauthor{\bsnm{Fr{\'e}nay}, \binits{B.}},
\bauthor{\bsnm{Verleysen}, \binits{M.}}:
\batitle{Classification in the presence of label noise: a survey}.
\bjtitle{IEEE transactions on neural networks and learning systems}
\bvolume{25}(\bissue{5}),
\bfpage{845}--\blpage{869}
(\byear{2013})
\end{barticle}
\endbibitem

\bibitem[\protect\citeauthoryear{Der~Kiureghian and Ditlevsen}{2009}]{der2009aleatory}
\begin{barticle}
\bauthor{\bsnm{Der~Kiureghian}, \binits{A.}},
\bauthor{\bsnm{Ditlevsen}, \binits{O.}}:
\batitle{Aleatory or epistemic? does it matter?}
\bjtitle{Structural safety}
\bvolume{31}(\bissue{2}),
\bfpage{105}--\blpage{112}
(\byear{2009})
\end{barticle}
\endbibitem

\bibitem[\protect\citeauthoryear{Seitzer et~al.}{2022}]{seitzer2022pitfalls}
\begin{botherref}
\oauthor{\bsnm{Seitzer}, \binits{M.}},
\oauthor{\bsnm{Tavakoli}, \binits{A.}},
\oauthor{\bsnm{Antic}, \binits{D.}},
\oauthor{\bsnm{Martius}, \binits{G.}}:
On the pitfalls of heteroscedastic uncertainty estimation with probabilistic neural networks.
arXiv preprint arXiv:2203.09168
(2022)
\end{botherref}
\endbibitem

\bibitem[\protect\citeauthoryear{Venkatesh and Thiagarajan}{2019}]{venkatesh2019heteroscedastic}
\begin{botherref}
\oauthor{\bsnm{Venkatesh}, \binits{B.}},
\oauthor{\bsnm{Thiagarajan}, \binits{J.J.}}:
Heteroscedastic calibration of uncertainty estimators in deep learning.
arXiv preprint arXiv:1910.14179
(2019)
\end{botherref}
\endbibitem

\bibitem[\protect\citeauthoryear{Zhang et~al.}{2018}]{zhang2018national}
\begin{barticle}
\bauthor{\bsnm{Zhang}, \binits{G.-Q.}},
\bauthor{\bsnm{Cui}, \binits{L.}},
\bauthor{\bsnm{Mueller}, \binits{R.}},
\bauthor{\bsnm{Tao}, \binits{S.}},
\bauthor{\bsnm{Kim}, \binits{M.}},
\bauthor{\bsnm{Rueschman}, \binits{M.}},
\bauthor{\bsnm{Mariani}, \binits{S.}},
\bauthor{\bsnm{Mobley}, \binits{D.}},
\bauthor{\bsnm{Redline}, \binits{S.}}:
\batitle{The national sleep research resource: towards a sleep data commons}.
\bjtitle{Journal of the American Medical Informatics Association}
\bvolume{25}(\bissue{10}),
\bfpage{1351}--\blpage{1358}
(\byear{2018})
\end{barticle}
\endbibitem

\bibitem[\protect\citeauthoryear{Posocco and Bonnefoy}{2021}]{posocco2021estimating}
\begin{bchapter}
\bauthor{\bsnm{Posocco}, \binits{N.}},
\bauthor{\bsnm{Bonnefoy}, \binits{A.}}:
\bctitle{Estimating expected calibration errors}.
In: \bbtitle{Artificial Neural Networks and Machine Learning--ICANN 2021: 30th International Conference on Artificial Neural Networks, Bratislava, Slovakia, September 14--17, 2021, Proceedings, Part IV 30},
pp. \bfpage{139}--\blpage{150}
(\byear{2021}).
\bcomment{Springer}
\end{bchapter}
\endbibitem

\bibitem[\protect\citeauthoryear{Garcia-Ceja et~al.}{2018}]{garcia2018depresjon}
\begin{bchapter}
\bauthor{\bsnm{Garcia-Ceja}, \binits{E.}},
\bauthor{\bsnm{Riegler}, \binits{M.}},
\bauthor{\bsnm{Jakobsen}, \binits{P.}},
\bauthor{\bsnm{T{\o}rresen}, \binits{J.}},
\bauthor{\bsnm{Nordgreen}, \binits{T.}},
\bauthor{\bsnm{Oedegaard}, \binits{K.J.}},
\bauthor{\bsnm{Fasmer}, \binits{O.B.}}:
\bctitle{Depresjon: a motor activity database of depression episodes in unipolar and bipolar patients}.
In: \bbtitle{Proceedings of the 9th ACM Multimedia Systems Conference},
pp. \bfpage{472}--\blpage{477}
(\byear{2018})
\end{bchapter}
\endbibitem

\bibitem[\protect\citeauthoryear{Jakobsen et~al.}{2020}]{jakobsen2020psykose}
\begin{bchapter}
\bauthor{\bsnm{Jakobsen}, \binits{P.}},
\bauthor{\bsnm{Garcia-Ceja}, \binits{E.}},
\bauthor{\bsnm{Stabell}, \binits{L.A.}},
\bauthor{\bsnm{Oedegaard}, \binits{K.J.}},
\bauthor{\bsnm{Berle}, \binits{J.O.}},
\bauthor{\bsnm{Thambawita}, \binits{V.}},
\bauthor{\bsnm{Hicks}, \binits{S.A.}},
\bauthor{\bsnm{Halvorsen}, \binits{P.}},
\bauthor{\bsnm{Fasmer}, \binits{O.B.}},
\bauthor{\bsnm{Riegler}, \binits{M.A.}}:
\bctitle{Psykose: A motor activity database of patients with schizophrenia}.
In: \bbtitle{2020 IEEE 33rd International Symposium on Computer-Based Medical Systems (CBMS)},
pp. \bfpage{303}--\blpage{308}
(\byear{2020}).
\bcomment{IEEE}
\end{bchapter}
\endbibitem

\bibitem[\protect\citeauthoryear{Bakker et~al.}{2023}]{bakker2023scoring}
\begin{barticle}
\bauthor{\bsnm{Bakker}, \binits{J.P.}},
\bauthor{\bsnm{Ross}, \binits{M.}},
\bauthor{\bsnm{Cerny}, \binits{A.}},
\bauthor{\bsnm{Vasko}, \binits{R.}},
\bauthor{\bsnm{Shaw}, \binits{E.}},
\bauthor{\bsnm{Kuna}, \binits{S.}},
\bauthor{\bsnm{Magalang}, \binits{U.J.}},
\bauthor{\bsnm{Punjabi}, \binits{N.M.}},
\bauthor{\bsnm{Anderer}, \binits{P.}}:
\batitle{Scoring sleep with artificial intelligence enables quantification of sleep stage ambiguity: hypnodensity based on multiple expert scorers and auto-scoring}.
\bjtitle{Sleep}
\bvolume{46}(\bissue{2}),
\bfpage{154}
(\byear{2023})
\end{barticle}
\endbibitem

\bibitem[\protect\citeauthoryear{Zheng et~al.}{2021}]{zheng2021uncertainty}
\begin{barticle}
\bauthor{\bsnm{Zheng}, \binits{R.}},
\bauthor{\bsnm{Zhang}, \binits{S.}},
\bauthor{\bsnm{Liu}, \binits{L.}},
\bauthor{\bsnm{Luo}, \binits{Y.}},
\bauthor{\bsnm{Sun}, \binits{M.}}:
\batitle{Uncertainty in bayesian deep label distribution learning}.
\bjtitle{Applied Soft Computing}
\bvolume{101},
\bfpage{107046}
(\byear{2021})
\end{barticle}
\endbibitem

\bibitem[\protect\citeauthoryear{Reamaroon et~al.}{2018}]{reamaroon2018accounting}
\begin{barticle}
\bauthor{\bsnm{Reamaroon}, \binits{N.}},
\bauthor{\bsnm{Sjoding}, \binits{M.W.}},
\bauthor{\bsnm{Lin}, \binits{K.}},
\bauthor{\bsnm{Iwashyna}, \binits{T.J.}},
\bauthor{\bsnm{Najarian}, \binits{K.}}:
\batitle{Accounting for label uncertainty in machine learning for detection of acute respiratory distress syndrome}.
\bjtitle{IEEE journal of biomedical and health informatics}
\bvolume{23}(\bissue{1}),
\bfpage{407}--\blpage{415}
(\byear{2018})
\end{barticle}
\endbibitem

\bibitem[\protect\citeauthoryear{Ju et~al.}{2022}]{ju2022improving}
\begin{barticle}
\bauthor{\bsnm{Ju}, \binits{L.}},
\bauthor{\bsnm{Wang}, \binits{X.}},
\bauthor{\bsnm{Wang}, \binits{L.}},
\bauthor{\bsnm{Mahapatra}, \binits{D.}},
\bauthor{\bsnm{Zhao}, \binits{X.}},
\bauthor{\bsnm{Zhou}, \binits{Q.}},
\bauthor{\bsnm{Liu}, \binits{T.}},
\bauthor{\bsnm{Ge}, \binits{Z.}}:
\batitle{Improving medical images classification with label noise using dual-uncertainty estimation}.
\bjtitle{IEEE transactions on medical imaging}
\bvolume{41}(\bissue{6}),
\bfpage{1533}--\blpage{1546}
(\byear{2022})
\end{barticle}
\endbibitem

\bibitem[\protect\citeauthoryear{Boeker}{2023}]{boeker2023ensemble}
\begin{botherref}
\oauthor{\bsnm{Boeker}, \binits{M.}}:
{Code for "A Novel Approach for Sleep Detection Using Weakly Supervised Learning"}.
GitHub.
\url{https://github.com/matthiasboeker/ensemble_sleep_detection}, Last accessed on 15th December 2023
(2023)
\end{botherref}
\endbibitem

\end{thebibliography}
\end{multicols}



\begin{appendices}

\section{Results of the Psykose and Depresjon Dataset}\label{sec:A1Data}
\begin{figure}[hbt!]
    \centering
    \includegraphics[width=1\linewidth]{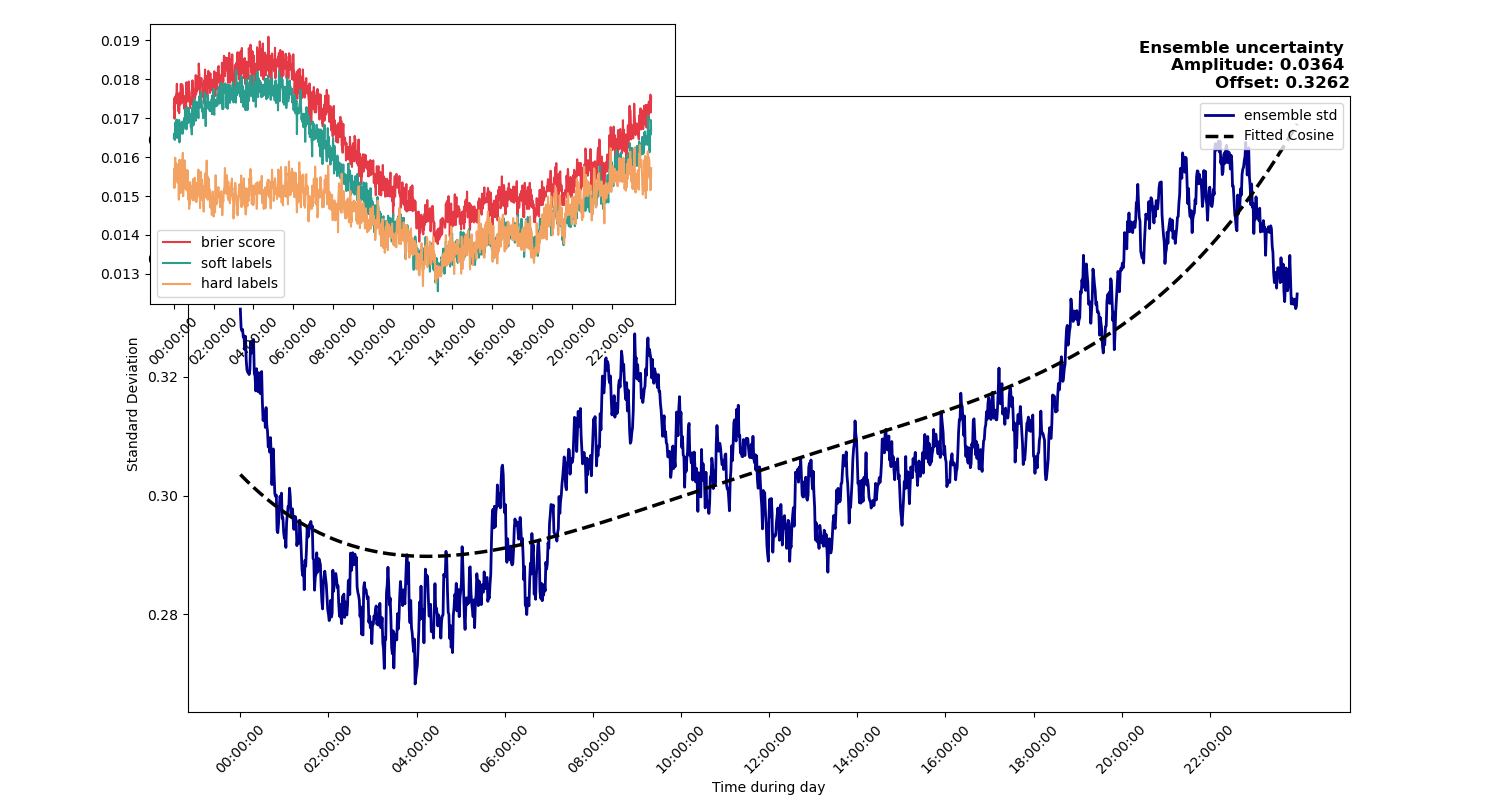}
    \caption{\textbf{The overall comparison of the time-dependent uncertainty of the ensemble model with the \gls{lstm} trained on three different loss functions, based on the Psykose and Depresjon dataset.}The figure compares the time dependent prediction uncertainty of the ensemble model with that of the \gls{lstm} trained with three different loss functions. The main diagram highlights the prediction uncertainty of the ensemble with a blue line, while the inset diagram provides a detailed view of the significantly smaller prediction uncertainties of the \gls{lstm} models for better visualisation. Additionally, the main plot features a dashed line, fitting a cubic curve with an amplitude of 0.0326 and an offset of 0.326. This cubic spline serves to model the heteroscedasticity of the prediction uncertainty.}
    \label{fig:mental_health_ensemble_uncertainty_comparision}
\end{figure}

\begin{figure}[hbt!]
    \centering
    \includegraphics[width=1\linewidth]{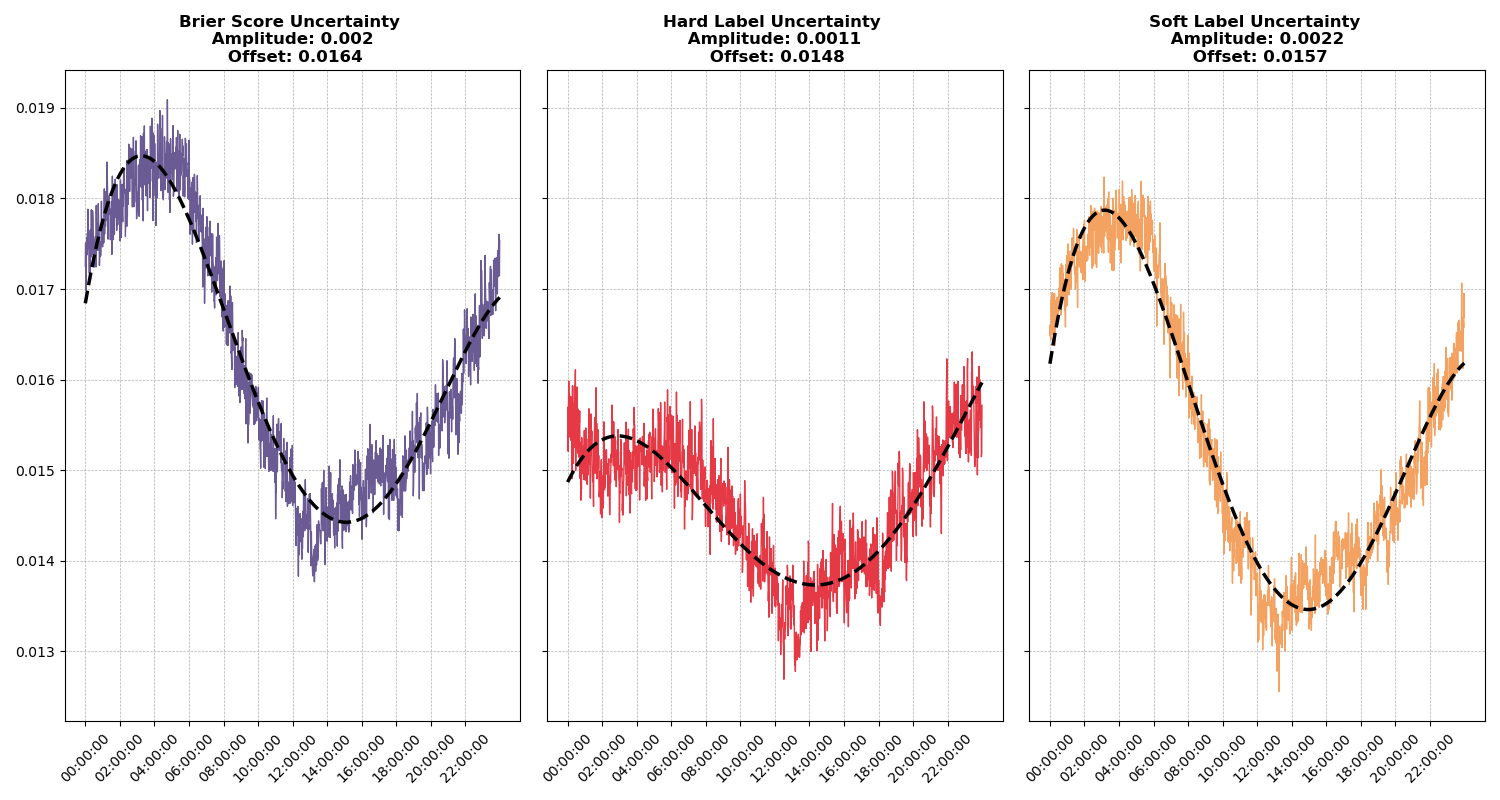}
    \caption{\textbf{The comparison of the time-dependent uncertainty of the loss function based on the Psykose and Depresjon dataset.} The figure compares the prediction uncertainty in size and shape of the \gls{lstm} trained with the three different loss functions. The figure is a detailed presentation of the inset plot of Figure~\ref{fig:mental_health_ensemble_uncertainty_comparision}. The dashed lines represent fitted cubic splines. The offset indicates the size of the uncertainty. The amplitude shows the structure.}
    \label{fig:mental_health_loss_function_uncertainty_comparision}
\end{figure}

In the scope of the \textit{Psykose} dataset, 22 participants recorded their motor activity with a wrist-worn actigraph device (Actiwatch, Cambridge Neurotechnology Ltd, England, model AW4) . The participants were 22 psychotic patients hospitalized at Haukeland University hospital. The participants recorded motor activity at a sampling rate of 32 Hz for an average duration of 12.6 days. In the post-processing, the activity was aggregated to 60 seconds epoch activity counts.
The \textit{Depresjon} dataset contains the motor activity over 12.6 days recording of 23 unipolar and bipolar depressed patients \cite{garcia2018depresjon}. The patients were also situated at the Haukeland University hospital. The experiment followed the same technical procedures as described for the \textit{Psykose} dataset. Additional to the 22 psychotic and 22 depressed patients, the datasets provide 33 control participants \cite{garcia2018depresjon, jakobsen2020psykose}. 
Neither dataset includes \gls{psg}, precluding validation of classification performance. Nevertheless, we can evaluate the prediction uncertainty in ensemble and dropout neural networks, focusing on heteroscedasticity. Figure~\ref{fig:mental_health_ensemble_uncertainty_comparision} shows similar to the results from the \gls{mesa} study, neural networks reduce the prediction uncertainty. Yet, distinct heteroscedastic patterns emerge across different loss functions. Figure~\ref{fig:mental_health_loss_function_uncertainty_comparision} further highlights that different loss functions exhibit heteroscedastic patterns. Specifically, the Brier score and soft cross-entropy function show significant temporal variations, reflecting increased prediction uncertainty during nighttime hours. Conversely, the hard cross-entropy demonstrates the most minimal heteroscedastic fluctuation.

\section{Derivation of soft cross-entropy loss as the MLE of binomial likelihood}\label{ap:proof}
We want to show that the gradient of the log-likelihood of the binominal distribution is equal to the soft cross-entropy loss function, in the case of binary classification. 
The cross-entropy loss for binary classification is given by
\begin{equation}
     H(p_t, y_t) = -y_t \ln(p_t) - (1 - y_t) \ln(1 - p_t)
\end{equation}
We consider the neural network with parameter $\theta$. $y_t$ is considered as the observed label and $p_t = \mathfrak{f} (\mathbf{x_t}, \theta)$ the output of the neural network. 
\begin{equation}
    H(\mathfrak{f} (\mathbf{x_t}, \theta), y_t) = -y_t \ln(\mathfrak{f} (\mathbf{x_t}, \theta)) - (1 - y_t) \ln(1 - \mathfrak{f} (\mathbf{x_t}, \theta))
\end{equation}
We derive the gradient of the cross-entropy loss with respect to $\theta$
\begin{equation}
    \nabla_{\theta} H(\mathfrak{f} (\mathbf{x_t}, \theta), y_t) = -y_t \nabla_{\theta} \ln(\mathfrak{f} (\mathbf{x_t}, \theta)) - (1 - y_t) \nabla_{\theta} \ln(1 - \mathfrak{f} (\mathbf{x_t}, \theta))
\end{equation}
In the hard label cross-entropy loss function, $y$ is a binary (0 or 1). However, in the case of soft cross-entropy loss, $y$ is replaced by $\hat{p}$, which refers to the average over the ensemble of labels.
\begin{equation}
\nabla_{\theta} H(\mathfrak{f} (\mathbf{x_t}, \theta), \frac{1}{k} \sum_{i=1}^k y_{it}) = -\left(\frac{1}{k} \sum_{i=1}^k y_{it} \right) \nabla_{\theta} \ln(\mathfrak{f} (\mathbf{x_t}, \theta)) - \left(1 - \frac{1}{k} \sum_{i=1}^k y_{it} \right) \nabla_{\theta} \ln(1 - \mathfrak{f} (\mathbf{x_t}, \theta))
\end{equation}
When we sum over the data $T$, we derive the gradient of the log-likelihood of the binomial distribution. 
\begin{equation}
    \sum_{t=1}^T \nabla_{\theta} H(\mathfrak{f} (\mathbf{x}, \theta), \frac{1}{k} \sum_{i=1}^k y_{it}) = \sum_{t=1}^T \left[\left(\frac{1}{k} \sum_{i=1}^n y_{it} \right) \nabla_{\theta} \ln \mathfrak{f} (\mathbf{x}, \theta) + \left(1- \frac{1}{k}\sum_{i=1}^k y_{it} \right) \nabla_{\theta}\ln (1 - \mathfrak{f} (\mathbf{x}, \theta)) \right]
\end{equation}

\end{appendices}


\end{document}